# Co-design of magnetic soft robots with large deformation and contacts via material point method and topology optimization

*Liwei Wang*[*]

*Department of Mechanical Engineering, Carnegie Mellon University, 5000 Forbes Avenue, Pittsburgh, 15213, PA, USA*

## Abstract

Magnetic soft robots embedded with hard magnetic particles enable untethered actuation via external magnetic fields, offering remote, rapid, and precise control, which is highly promising for biomedical applications. However, designing such systems is challenging due to the complex interplay of magneto-elastic dynamics, large deformation, solid contacts, time-varying stimuli, and posture-dependent loading. While these challenges are particularly prominent in magnetic soft robots, they are also fundamental and common to responsive materials in general. As a result, most existing research relies on heuristics and trial-and-error methods or focuses on the independent design of stimuli or structures under static conditions. We propose a topology optimization framework for magnetic soft robots that simultaneously designs structures, location-specific material magnetization, and time-varying magnetic stimuli, accounting for large deformations, dynamic motion, and solid contacts. This is achieved by integrating generalized topology optimization with the magneto-elastic material point method, which supports GPU-accelerated parallel simulations and automatic differentiation for sensitivity analysis. We applied this framework to design magnetic robots for various tasks, including multi-task shape morphing and locomotion, in both 2D and 3D. The method autonomously generates optimized robotic systems to achieve target behaviors without requiring human heuristics. Despite the nonlinear physics and large design space, it demonstrates high computational efficiency, completing all cases within minutes. While we focus on magnetic soft robots in this study, the proposed method can be readily extended to other co-design problems involving stimuli-responsive materials with large deformations and complex dynamics. The framework provides a computational foundation for the autonomous co-design of active soft materials in applications such as metasurfaces, drug delivery, and minimally invasive procedures.

[*] Corresponding author: liweiw@andrew.cmu.edu

## 1. Introduction

Soft robots are made from flexible and stretchable materials with inherent mechanical compliance, offering greater degrees of freedom, adaptability, and biocompatibility compared to conventional rigid robots [1, 2]. Many soft robots rely on tethered actuators, such as pneumatic or tendon-driven systems, which limit their mobility and scalability with physical constraints like tubes and wires. Recent advances in materials and manufacturing have enabled the creation of untethered soft robots using stimuli-responsive materials [3-5]. These materials change properties or generate internal forces in response to external stimuli, driving the robots' deformation and movement. One particularly promising type of untethered soft robot is the magnetic soft robot [6-8], as shown in Fig. 1. It is made from a soft polymer matrix embedded with hard-magnetic particles, which have high coercivity and retain strong magnetization after being magnetized. These materials, known as hard-magnetic soft materials, interact with external magnetic fields to generate forces or torques, driving the overall deformation and motion of the soft matrix [9-12]. By altering the magnitude and orientation of the external magnetic field, we can remotely control these robots with high precision and speed. This precise, rapid, remote, and directional actuation distinguishes magnetic soft robots from those relying on less controllable stimuli like temperature [13], humidity [14], or pH [15]. They have shown great potential in applications such as drug delivery [16], minimally invasive procedures [8, 17-19], soft electronics [20, 21], shape morphing [22, 23], and mechanical computing [24], across scales from micrometers to centimeters.

While magnetic soft robots offer enhanced control flexibility and greater freedom of movement, they also introduce significant complexity to the design process. Specifically, their behavior is governed by a complex interplay of material properties (magnetization distribution), structural configuration, and applied stimuli, all coupled through magneto-elastic interactions involving large deformations [6, 9, 11]. For dynamic applications, time-dependent factors add another layer of complexity. In these scenarios, magnetic robots experience time-varying and posture-dependent magnetic forces throughout their motion. Numerous pioneering theoretical and experimental studies have explored the dynamic behavior of related systems, such as snap-through and buckling in bistable structures [25, 26], as well as vibration and band-gap phenomena in laminates and actuators [27-30]. However, the complexity of magnetic robots is further heightened by their continuous formation and breakage of contact with the environment, leading to time-varying boundary conditions. Though finite element models (FEM) have been used for forward simulations of given robot designs, they are computationally expensive and difficult to implement [9]. If forward simulation is challenging, inverse design is even more demanding, which requires determining high-dimensional design variables to achieve specific tasks. This challenge is especially pronounced when co-designing is necessary to orchestrate the structure, spatially varying material properties, and time-dependent stimuli.

As a result, while remarkable applications have been demonstrated with existing magnetic soft robots [6, 7, 23], most have been developed through heuristic or trial-and-error approaches, which lack the flexibility to accommodate complex or customized requirements [31]. To address these challenges, some research has explored exhaustive parameter sweeps [18, 32] or gradient-free optimization techniques [33-35], to explore the design space in a more systematic way. While these methods have shown promise for static designs with a small number of parameters, they are generally inefficient and do not scale well for more intricate dynamic designs, as they require numerous design evaluations.

To improve efficiency, some studies have incorporated machine learning models as surrogates for expensive simulations within the iterative optimization [36]. This surrogate-based framework has also been applied to the design of dynamic jumping robots, using simulated data generated from a mass-spring lattice model incorporating evolutionary optimization [17]. Reinforcement learning has also been used to guide control strategies, though it has not yet considered structural or materials design [37, 38]. Iteration-free machine learning methods that map targets directly to design variables have also emerged [39]. However, these methods often require large datasets/samples and are limited to the conditions included in the data, making them less adaptable to new tasks or complex cases. Moreover, most existing studies optimize either structure or material properties alone, limiting the functional potential of magnetic soft robots.

To enhance design flexibility, topology optimization (TO) methods have been introduced for hard-magnetic soft materials. TO can consider free-form shapes [40], topological changes, multi-physics coupling [41-44] and large deformations [45-49], providing greater adaptability compared to parametric methods. For example, level set methods combined with conformal mapping have been used to design active structures on free-form surfaces, optimizing magnetic soft material distribution with homogeneous magnetization [50]. Density-based topology optimization has been employed to design composite structures that tune band gap behaviors under magnetic fields [51, 52]. More recently, density-based methods have been further extended to co-design materials, structures, and magnetic field orientations for metamaterials, metasurfaces, and shape morphing [53-56]. While these co-design breakthroughs have significantly advanced design flexibility for magnetic soft robots, they are still restricted to static or quasi-static designs with fixed boundary conditions. They focus primarily on shape-morphing applications rather than achieving dynamic motions with nonlinear contacts, which are crucial for real-world robotic applications. Recent research has enabled the gradient-based design of control strategies for magnetic soft robots with dynamic contacts. However, it has primarily focused on stimulus design, without incorporating material and structural design [57].

Therefore, a unified inverse design framework for magnetic soft robots is still missing—one that can (1) co-design spatially varying materials, structures, and time-dependent magnetic stimuli, (2) account for dynamic motion, (3) consider the strong coupling between actuation and posture in motion, and (4) handle large deformations and evolving contacts [31]. This gap extends beyond magnetic soft robots, highlighting a common bottleneck in the inverse design of responsive material systems. It stems from two key challenges: how to simulate complex dynamic solids accurately and efficiently, and how to evaluate the sensitivity of design variables to guide the optimization. Most current designs rely on Lagrangian FEM, which struggles with large deformations and dynamic contacts, leading to convergence issues, high computational costs, and challenges in obtaining sensitivity values for optimization. These limitations make it challenging for FEM-based methods to address the complexities of designing various responsive material systems, with magnetic soft robot design as a representative example.

The Material Point Method (MPM) has recently emerged as a promising alternative to FEM [58, 59]. MPM uses Lagrangian particles to represent materials undergoing large deformations, while a fixed Eulerian mesh effectively handles contacts. Its parallelizable nature makes it suitable for GPU acceleration, significantly improving computational efficiency [60]. MPM has demonstrated accuracy and stability in simulating complex interactions involving solids, fluids, and granular materials, handling large deformations, multiphase coupling [61], and solid contacts [62-66]. More recently, MPM has also been used effectively to simulate magnetic soft robots in both quasi-static and dynamic settings, demonstrating high efficiency and strong agreement with experimental results [67, 68]. Additionally, MPM can be implemented with automatic differentiation (AD) [69], allowing direct computation of sensitivity values for gradient-based optimization, as shown in recent work on topology optimization for tethered pneumatic soft robots [70-72].

In this study, we propose a unified inverse design framework that integrates magneto-elastic MPM

simulations with topology optimization. It enables the co-design of materials, structures, and stimuli for untethered magnetic soft robots, accounting for large deformations and dynamic contacts. As shown in Fig. 1, we use particles to represent the robot body to be designed. For materials at each particle region, we design the local remanent magnetization (both orientation and magnitude), forming a vector field $\widetilde{\mathbf{B}}_r$. For the structural design, we use a binary variable to represent void or solid state for each particle, and relax it into a continuous scalar field $\boldsymbol{\varphi}$ with penalization and smoothed projection for nonbinary values. For stimuli, we design both orientation and time-varying magnitude. By leveraging an interpolated magneto-elastic Helmholtz energy function, we unify materials, structure, and stimuli to create a comprehensive design space for magnetic robotic systems.

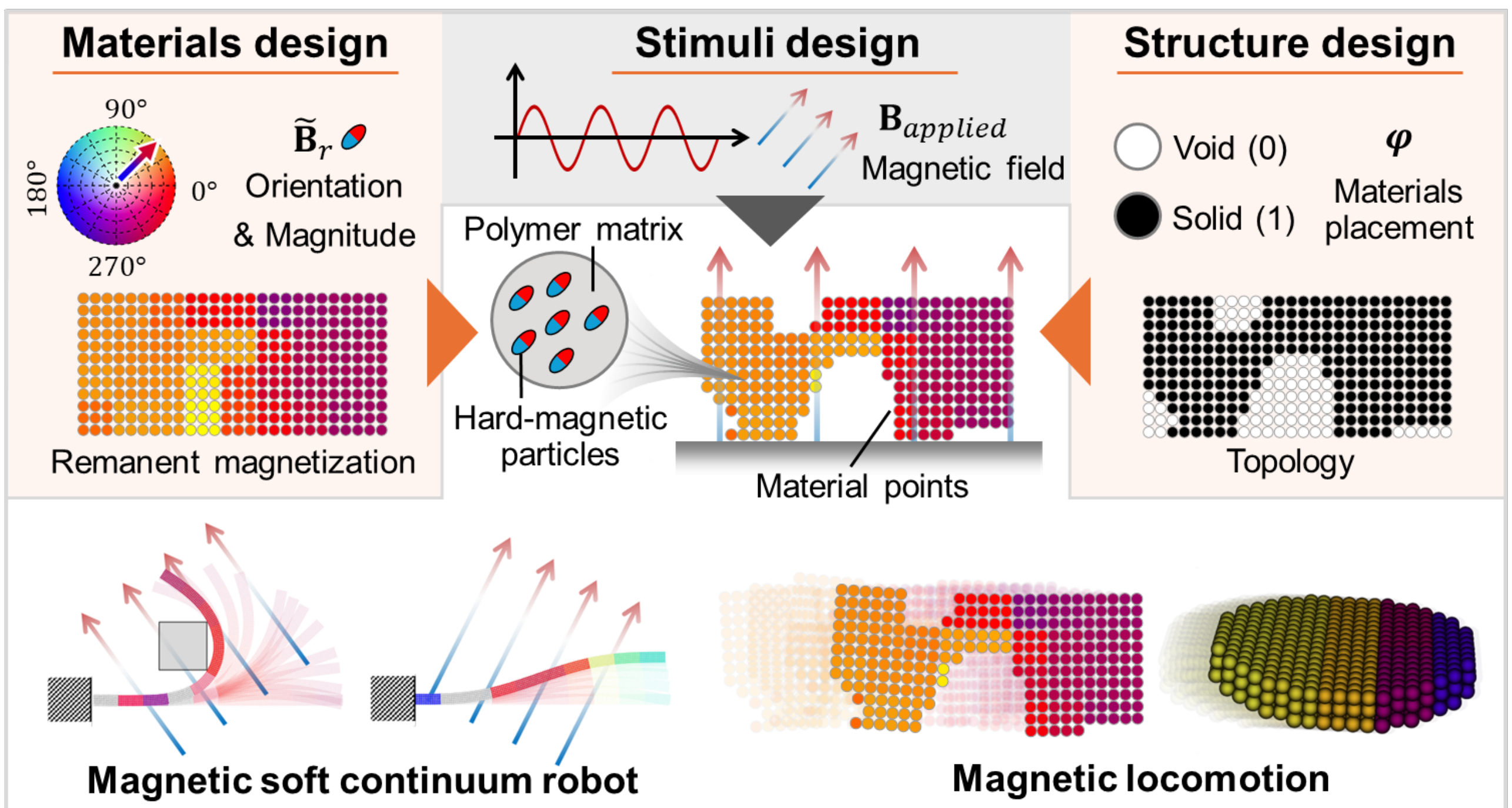

Fig. 1. Overview of the proposed co-design framework and its applications.

We then use the MPM to simulate the entire dynamic process, accounting for large deformations and dynamic contacts. The MPM is implemented with GPU-enabled parallel computing and supports automatic differentiation (AD) to compute sensitivity [60, 69]. These two techniques, which have previously advanced AI and machine learning, are now leveraged to accelerate physical simulations and facilitate gradient-based topology optimization. This approach enables efficient co-design of structures, heterogeneous materials, and time-varying stimuli, providing significant design flexibility to adapt to customized needs and specific tasks.

As a demonstration, we apply the proposed framework to two common applications of magnetic soft robots: (1) a thread-like magnetic soft continuum robot for minimally invasive surgery, achieving multi-task quasi-static shape morphing, and (2) magnetic soft locomotion, moving in a specified direction under cyclic magnetic excitations in both 2D and 3D cases. In all scenarios, the design is autonomously generated without human intervention. The framework demonstrates exceptional efficiency and efficacy, completing the design process within minutes despite nonlinearities, complex physics, and contact dynamics. The proposed framework has the potential to significantly accelerate the design of various soft robotic applications, including drug delivery [16], minimally invasive surgery [17], and soft electronics.

In summary, the main contributions of our work include:

- We propose a unified and differentiable design representation that accommodates diverse spatiotemporal entities in magnetic soft robots, including spatially varying scalar and vector fields within the robot body, as well as time-dependent, oriented stimuli.
- We couple this high-dimensional spatiotemporal representation with the energy formulation and physical properties in magneto-elastic MPM, enabling efficient GPU-parallelized computation and automatic sensitivity analysis.
- We develop an integrated design framework with general optimization formulations that support the co-design of materials, structure, and stimuli for both static/quasi-static and dynamic applications, accommodating posture-dependent magnetic actuation, large deformations, and evolving contacts.

- Through comparative studies, we reveal the distinct roles and interactions among materials, structure, and stimuli, identifying scenarios where full co-design is essential versus cases where partial design suffices. These insights offer valuable guidance for future soft robotic design strategies beyond magnetic systems.

The remainder of the paper is organized as follows. Section 2 introduces the magneto-elastic constitutive model and its integration into MPM. Section 3 details the proposed co-design framework based on the MPM. In Section 4, we demonstrate the proposed design framework through various case studies, covering both quasi-static and dynamic scenarios in 2D and 3D. Conclusions and discussions are presented in Section 5.

## 2. Differentiable dynamic simulation of magnetic soft robots

In this section, we present the constitutive model for hard-magnetic soft materials and integrate it into the MPM to enable dynamic simulation under an external magnetic field. We highlight the handling of dynamic contacts and the automatic differentiation of MPM to extract gradient information, which forms the foundation for the co-design framework discussed in later sections.

### 2.1 Constitutive model and governing equations of hard-magnetic soft material

Consider a continuum solid made of hard-magnetic soft materials occupying a domain $\Omega_0$ in the reference state, with material particles labeled by their position vectors $\mathbf{X}$. After deformation, it goes to the current state $\Omega$ with particle position vectors $\mathbf{x}$, which corresponds to a smooth deformation map $\mathbf{x} = \chi(\mathbf{X})$. The deformation gradient tensor is defined as $\mathbf{F} = \frac{\partial \mathbf{x}}{\partial \mathbf{X}}$ with its Jacobian $J = \det \mathbf{F}$. The hard-magnetic soft materials are magnetized under a strong magnetic field, saturating the embedded particles and resulting in a residual magnetic flux density. We denote the residual magnetic flux density in the reference and current states to be $\widetilde{\mathbf{B}}_r$ and $\mathbf{B}_r$, respectively. The relation between them can be given as [7,

9, 11]:

$$\mathbf{B}_r = J^{-1}\mathbf{F}\widetilde{\mathbf{B}}_r, \qquad \widetilde{\mathbf{B}}_r = J\mathbf{F}^{-1}\mathbf{B}_r\,. \tag{1}$$

**Remark 1**

As shown in Eq. 1, the residual magnetic flux densities at different states are linked through the deformation gradient $\mathbf{F}$, known as the $\mathbf{F}$-based model [9]. This contrasts with the more recent $\mathbf{R}$-based model, which uses only the rotational part $\mathbf{R}$ and assumes that remanent magnetization changes with rotation but not with stretching [25, 26, 73-75]. Our study focuses on slender beams, plates, and some 2D/3D locomotion dominated by bending or rigid body motion, where $\mathbf{F}$ and $\mathbf{R}$ are nearly equivalent. In such cases, we choose to adopt $\mathbf{F}$-based model, which offers a simpler formulation with sufficient accuracy, as supported by prior experiments [26, 35, 73, 76-80]. However, we note that, $\mathbf{F}$-based model is an approximation for bending-dominated cases while $\mathbf{R}$-based model is proven to be more consistent with the experiments across different cases [25, 26, 73-75]. In cases involving significant stretching, such as pre-stretched materials or deformation under nonuniform magnetic fields, the $\mathbf{R}$-based model is more appropriate and has been shown to yield better agreement with experimental results. For more details on these two models, we refer the reader to recent studies in the literature [25, 26, 73-75].

Due to the high coercivity of hard magnetic particles, the material maintains residual magnetic flux under a wide range of external magnetic fields $\mathbf{H}$ below the coercive field strength. We consider ideal hard-magnetic soft materials here [9], which assumes that 1) the permeability is close to that of a vacuum or air $\mu_0$, 2) the residual magnetic flux density $\widetilde{\mathbf{B}}_r$ remains constant within the working range of external actuation, and 3) its magnetic flux density $\mathbf{B}$ has a linear relation with $\mathbf{H}$ in this range:

$$\mathbf{H} = \frac{1}{\mu_0}(\mathbf{B} - \mathbf{B}_r). \tag{2}$$

Note that the ideal hard magnetic soft material assumption yields good agreement with experiments only when the applied magnetic fields are small, i.e., significantly lower than the coercivity of the embedded magnetic particles, with negligible magnetic dissipation effects [9, 17, 25]. Since our primary focus is on the optimization framework, we adopt this assumption for simplicity, as is common in many practical applications [6, 9, 11]. For example, hard magnetic NdFeB particles, widely used in 3D printing, can exhibit coercivity exceeding 1000 kA/m [6, 81]. In our design cases, the applied magnetic fields are all well below this threshold. We also note that, by adopting the ideal hard magnetic soft material assumption, our model neglects the effects of the surrounding air and magnetic source. This simplification has been widely used in previous studies and has demonstrated reasonable accuracy for material systems similar to those in this work [23, 33, 35, 36, 50, 53-56, 79]. However, in scenarios involving very strong magnetic fields, magnetic dissipation or coupling effects, the above ideal hard magnetic soft materials assumptions may no longer hold and fully coupled dissipative models would be needed. Further details can be found in [73, 75, 82].

The nominal Helmholtz free energy function for ideal hard-magnetic soft materials, which represents the Helmholtz free energy per unit reference volume, is expressed as [9]:

$$\widetilde{W}\left(\mathbf{F}, \widetilde{\mathbf{B}}\right) = \widetilde{W}_e(\mathbf{F}) + \widetilde{W}_m\left(\mathbf{F}, \widetilde{\mathbf{B}}\right), \tag{3}$$

where $\widetilde{W}_e(\mathbf{F})$ is the elastic energy as a function of $\mathbf{F}$, and $\widetilde{W}_m\left(\mathbf{F}, \widetilde{\mathbf{B}}\right)$ is the magnetic potential energy as a function of $\mathbf{F}$ and magnetic flux density $\widetilde{\mathbf{B}}$ in the reference state, given an external magnetic field $\mathbf{H}$ and residual magnetic flux density $\widetilde{\mathbf{B}}_r$. For elastic energy, we use the Neo-Hookean model

$$\widetilde{W}_e = \frac{G}{2}\left(J^{-\frac{2}{3}} I_1 - 3\right) + \frac{K}{2}(J-1)^2, \tag{4}$$

where $G$ is the shear modulus, $K$ is the bulk modulus, and $I_1 = \mathrm{trace}(\mathbf{F}^T\mathbf{F})$. For magnetic potential energy, we assume that the applied magnetic field $\mathbf{H}$ is uniform, corresponding to an applied magnetic flux density $\mathbf{B}_{applied} = \mu_0 \mathbf{H}$. Under the ideal hard-magnetic soft materials assumption, the magnetic potential is given as

$$\widetilde{W}_m = -\frac{1}{\mu_0} \mathbf{F}\widetilde{\mathbf{B}}_r \cdot \mathbf{B}_{applied}. \tag{5}$$

Note that in this study, we assume the magnetic interactions between components are negligible during actuation. This assumption is made based on the relatively low remanent magnetization in our designs and the fact that magnetic interaction forces decay quadratically with relative distance [83]. Similar assumptions have been adopted and validated in previous studies with comparable configurations [17, 18, 35, 54, 55, 79]. Nevertheless, for systems with significantly stronger magnetization, it would be important to account for interaction energy in the modeling [83]. With the defined elastic and magnetic energy, the total Helmholtz free energy per unit reference volume is

$$\widetilde{W} = \frac{G}{2}\left(J^{-\frac{2}{3}} I_1 - 3\right) + \frac{K}{2}(J-1)^2 - \frac{1}{\mu_0} \mathbf{F}\widetilde{\mathbf{B}}_r \cdot \mathbf{B}_{applied}, \tag{6}$$

from which we can derive the first Piola-Kirchhoff stress tensor as

$$\mathbf{P} = \frac{\partial \widetilde{W}}{\partial \mathbf{F}} = GJ^{-\frac{2}{3}}\left(\mathbf{F} - \frac{I_1}{3}\mathbf{F}^{-T}\right) + KJ(J-1)\mathbf{F}^{-T} - \frac{1}{\mu_0}\mathbf{B}_{applied} \otimes \widetilde{\mathbf{B}}_r, \tag{7}$$

where $\otimes$ is the dyadic product. The Cauchy stress is then given by

$$\boldsymbol{\sigma} = \frac{1}{J}\mathbf{P}\mathbf{F}^T = GJ^{-\frac{5}{3}}\left(\mathbf{F}\mathbf{F}^T - \frac{I_1}{3}\mathbf{I}\right) + KJ(J-1)\mathbf{I} - \frac{1}{\mu_0 J}\mathbf{B}_{applied} \otimes \mathbf{F}\widetilde{\mathbf{B}}_r. \tag{8}$$

Based on Eq. 1, we can further derive

$$\boldsymbol{\sigma} = GJ^{-\frac{5}{3}}\left(\mathbf{F}\mathbf{F}^T - \frac{I_1}{3}\mathbf{I}\right) + KJ(J-1)\mathbf{I} - \frac{1}{\mu_0}\mathbf{B}_{applied} \otimes \mathbf{B}_r. \tag{9}$$

The first two terms arise from elastic deformation, while the last term represents the stress induced by the external magnetic field. Together, they describe the interaction between magnetism and elasticity. Based on this, we can express the corresponding body torque $\mathbf{m}$ generated by the magnetized materials under the applied magnetic field as

$$\mathbf{m} = \frac{1}{\mu_0}\mathbf{B}_r \times \mathbf{B}_{applied}, \tag{10}$$

which represents the torque per unit volume in the deformed configuration. Notably, Eq. 10 indicates that this magnetic body torque depends on the cross-product of the residual magnetic flux $\mathbf{B}_r$ and the external magnetic field $\mathbf{B}_{applied}$. As a result, the magnetic body torque reaches its maximum when the two vectors are perpendicular and becomes zero when they are parallel.

**Remark 2**

We note that the total Cauchy stress in Eq. 9 is asymmetric. As explained in the seminal works [7, 9] that established the magnetoelastic theory used here, this asymmetry originates from magnetic body torques generated by embedded magnetized particles subjected to externally applied magnetic fields. Using the stress and body torque expressions in Eqs. 9-10, it has been shown that the conservation of angular momentum is satisfied [9]. Recently, there has been active discussion on this asymmetric expression, with several studies proposing alternative energy formulations that yield symmetric stress tensors [73, 75]. However, resolving this theoretical discussion lies beyond the scope of the present work. The proposed optimization framework can, in principle, accommodate various energy formulations. Accordingly, we follow established precedents with similar settings to use this theory, which has been experimentally validated under conditions comparable to those considered here [17, 23, 32-36, 50, 53-56, 77-79, 84]. Readers interested in the ongoing theoretical discussions are referred to the most recent literature [25, 73, 75, 76, 85].

The governing equations can be expressed in the Lagrangian view (reference state $\mathbf{X} \in \Omega_0$):

$$\rho_0(\mathbf{X}, \mathrm{t}) = J\rho_0(\mathbf{X}, 0), \qquad \text{Conservation of mass} \tag{11}$$

$$\nabla_0 \cdot \mathbf{P} + \mathbf{f}_0 = \rho_0 \frac{\partial \mathbf{V}}{\partial \mathrm{t}}, \qquad \text{Conservation of momentum} \tag{12}$$

and in the Eulerian view (current state $\mathbf{x} \in \Omega_0$):

$$\frac{D\rho}{Dt} + \rho\nabla \cdot \mathbf{v} = 0, \qquad \text{Conservation of mass} \tag{13}$$

$$\nabla \cdot \boldsymbol{\sigma} + \mathbf{f} = \rho \frac{D\mathbf{v}}{Dt}, \qquad \text{Conservation of momentum} \tag{14}$$

where $\nabla$, $\mathbf{f}(\mathbf{x}, \mathrm{t})$ and $\rho(\mathbf{x}, \mathrm{t})$ represent the gradient vector, body force, and density in the current state, respectively. Their counterparts in the reference state are denoted with a subscript 0. The material velocity field is defined as $\mathbf{v}(\mathbf{X}, t) = \mathbf{V}(\mathbf{x}, t)$ , and $D(\cdot)/Dt = \partial(\cdot)/\partial t + \mathbf{v} \cdot \nabla(\cdot)$ denotes the material time derivative. It is important to note that the body force $\mathbf{f}$ here does not include the forces induced by an external magnetic field, as the magnetic contribution is already accounted for in the stress term.

### 2.2 Magneto-elastic material point method

As demonstrated in the derivation of constitutive models and governing equations, a Lagrangian perspective is often more natural in solid continuum mechanics for defining state variables and deriving governing equations. Consequently, FEM for solids typically employs a Lagrangian mesh to represent the material body and approximate the governing equations (Eqs. 11–12). However, a purely Lagrangian mesh can suffer from severe distortion and numerical issues under large deformations, while also posing challenges in handling dynamic contact.

In contrast, the MPM is a hybrid Lagrangian/Eulerian approach (Fig. 2). It represents the body using material particles, tracking their states within a Lagrangian framework while computing derivatives and solving the governing equations on a fixed Eulerian mesh (Eqs. 13–14). This approach preserves the rigor of the Lagrangian method in deriving weak forms while leveraging the Eulerian mesh to effectively handle large deformations and collisions. In this study, we integrate the constitutive model and governing equations of hard-magnetic soft materials with Moving Least Squares MPM (MLS-MPM) [86] to simulate the responsive behaviors of hard-magnetic soft robots under actuation. Here, we provide a brief overview of the model; for more details, readers are referred to previous studies [58, 67, 86].

As shown in Fig. 2A, we represent the material in the simulation using a set of particles, each assigned a defined mass and volume. It is important to note that MPM particles are not physical entities, nor are they inherently circular or spherical. Instead, each MPM particle represents a continuous portion of the material—a subset of the entire simulated domain used solely for computational purposes. While we use circles or spheres to visualize MPM simulation results, they serve only for illustrative purposes. We typically initialize the particles by distributing them uniformly within the material body. Each particle $p$ is assigned a set of state variables, including volume $v_p$, mass $m_p$, position $\mathbf{x}_p(t)$, velocity $\mathbf{v}_p(t)$, an affine velocity matrix $\mathbf{C}_p(t)$, and a deformation gradient $\mathbf{F}_p(t)$ at the time $t$. Additionally, material properties such as residual magnetization ${}^p\widetilde{\mathbf{B}}_r$, and other constitutive parameters are stored within these particles.

The particles coexist with a background Eulerian mesh, which defines the fixed spatial domain where the material deforms and moves. We assume the mesh has an element size of $\Delta\mathrm{x}$. Each node $i$ stores the node position $\widetilde{\boldsymbol{x}}_i$, grid mass $\widetilde{m}_i(t)$ and grid velocity $\widetilde{\boldsymbol{v}}_i(t)$. Note that the grid itself does not deform ($\widetilde{\boldsymbol{x}}_i$ remains constant), while the grid mass and velocity are updated over time. MPM simulates iteratively, with each iteration exchanging information between the Lagrangian particles and the Eulerian mesh, following the steps outlined below:

1) **Particle to grid transfer (P2G).** This step transfers particle mass and momentum to the grid nodes of the background mesh. In the MLS-MPM [86], this transfer is achieved using a quadratic B-spline kernel distributed over a neighborhood of nodes, as shown in Fig. 2B. Specifically, for each node $i$, we can obtain the grid mass and momentum as

$$\widetilde{m}_i(t) = \sum_{p \in \mathbf{S}_i} N(\mathbf{x}_p(t) - \widetilde{\boldsymbol{x}}_i)\, m_p, \tag{15}$$

$$\widetilde{m}_i(t)\widetilde{\mathbf{v}}_i(t) = \sum_{p \in \mathbf{S}_i} N(\mathbf{x}_p(t) - \widetilde{\boldsymbol{x}}_i) \cdot \left[ m_p \mathbf{v}_p(t) + \left( m_p \mathbf{C}_p(t) + \frac{4}{\Delta\mathrm{x}^2} v_p \mathbf{P}_p \mathbf{F}_p^T(t) \right) \left( \mathbf{x}_p(t) - \widetilde{\boldsymbol{x}}_i \right) \right], \tag{16}$$

where $\mathbf{S}_i$ is the predefined neighboring region for node $i$, $N(\cdot)$ is the quadratic interpolation shape function, and $\mathbf{P}_p$ is the particle magneto-elastic stress computed using the particle deformation gradient $\mathbf{F}_p(t)$ via Eq. 7. This transfer essentially assigns distance-decaying weights to particles within the neighborhood and aggregates particle properties using a weighted sum to compute the grid mass and momentum of the node. In this study, we define the neighborhood $\mathbf{S}_i$ of a point as regions within a Chebyshev distance of element size $\Delta$x.

2) **Grid operation.** After the transfer from particle to grid, as shown in Fig. 2C, we compute the grid velocity as

$$\tilde{\mathbf{v}}_i(t) = \frac{1}{\tilde{m}_i(t)}(\tilde{m}_i(t)\tilde{\mathbf{v}}_i(t) + \mathbf{f}_i(t)\Delta t), \tag{17}$$

where $\Delta t$ is the time step and $\mathbf{f}_i(t)$ is the external force, including the damping force and gravity

$$\mathbf{f}_i(t) = -c\tilde{m}_i(t)\tilde{\mathbf{v}}_i(t) + \tilde{m}_i(t)\boldsymbol{g}, \tag{18}$$

where $c$ is the damping constant and $\boldsymbol{g}$ is the gravity constant. We then modify the velocity to impose boundary constraints and handle contacts, as shown in the right panel of Fig. 2C. Consider solid boundaries $\partial\Omega_b$, such as the ground and obstacles, with its outward-pointing normal vector denoted as $\mathbf{n}_b$. For nodes inside the solid boundaries $\partial\Omega_b$, we set their velocity to the constrained values $\mathbf{v}_b(t)$. These solid boundaries can be either moving or static ($\mathbf{v}_b(t) \equiv 0$). We first consider the case where the material is not directly attached to the boundary, allowing for relative movement between the material and the boundary. In this case, for nodes on $\partial\Omega_b$, i.e., $\tilde{\boldsymbol{x}}_i \subset \partial\Omega_b$, if grid velocity points towards the boundary, i.e., $(\tilde{\mathbf{v}}_i(t) - \mathbf{v}_b(t)) \cdot \mathbf{n}_b < 0$, we adjusted the grid velocity using:

$$\tilde{\mathbf{v}}_i(t) = \mathbf{v}_b(t) + max\left(0, \left|\tilde{\mathbf{v}}_{ib,\parallel}(t)\right| - \mu_f\left|\tilde{\mathbf{v}}_{ib,\perp}(t)\right|\right) \cdot \frac{\tilde{\mathbf{v}}_{ib,\parallel}}{\left|\tilde{\mathbf{v}}_{ib,\parallel}(t)\right|}, \tag{19}$$

where $\tilde{\mathbf{v}}_{ib}(t) = \tilde{\mathbf{v}}_i(t) - \mathbf{v}_b(t)$ is the relative velocity, $\tilde{\mathbf{v}}_{ib,\perp}(t)$ and $\tilde{\mathbf{v}}_{ib,\parallel}(t)$ are its normal and tangent components, respectively, and $\mu_f$ is the dynamic friction coefficient. By doing this, we essentially set the normal component of $\tilde{\mathbf{v}}_i(t)$ equal to that of $\mathbf{v}_b(t)$, while the tangential component is reduced according to Coulomb friction [70]. For solid boundaries with a very large friction coefficient, we assign non-slip correction instead, i.e., $\tilde{\mathbf{v}}_i(t) = \mathbf{v}_b(t)$, if grid velocity points towards the boundary. Finally, for boundaries directly attached to the material, we simply keep the $\tilde{\mathbf{v}}_i(t) = \mathbf{v}_b(t)$, regardless of the relative velocity.

3) **Grid to particle transfer (G2P).** As shown in Fig. 2D, the updated grid velocity is transferred back to the particles using the same quadratic B-spline kernel as in P2G. This step updates the particle velocity and affine velocity as follows:

$$\mathbf{v}_p(t+\Delta t) = \sum_{i \in \mathbf{S}_p} N(\mathbf{x}_p(t) - \tilde{\boldsymbol{x}}_i)\, \tilde{\mathbf{v}}_i(t), \tag{20}$$

$$\mathbf{C}_p(t+\Delta t) = \frac{4}{\Delta \mathrm{x}^2} \sum_{i \in \mathbf{S}_p} N(\mathbf{x}_p(t) - \tilde{\boldsymbol{x}}_i)\, \tilde{\mathbf{v}}_i(t) (\mathbf{x}_p(t) - \tilde{\boldsymbol{x}}_i)^{\mathrm{T}}, \tag{21}$$

where $\mathbf{S}_p$ is the neighboring region defined in P2G for the grid node closest to the particle $p$. The particle position and deformation gradient are updated by

$$\mathbf{x}_p(t+\Delta t) = \mathbf{x}_p(t) + \mathbf{v}_p(t+\Delta t)\Delta t, \tag{22}$$

$$\mathbf{F}_p(t+\Delta t) = \big(\mathbf{I} + \mathbf{C}_p(t+\Delta t)\Delta t\big)\mathbf{F}_p(t). \tag{23}$$

In the Lagrangian view, this step corresponds to a deformation of the materials represented by the particles, as shown in Fig. 2E.

The MPM process iterates until the specified simulation time T is reached. Throughout the entire process, the mesh remains fixed, acting as a surrogate to transfer information between particles from one step to the next, addressing the contacts and boundary constraints. Meanwhile, the particles are not constrained by the grid and can move freely across elements, enabling the simulation of large material deformations.

Intuitively, we can consider the Lagrangian particles as adaptive integration points for the Eulerian mesh, mitigating numerical issues associated with distorted Lagrangian FEM while ensuring accuracy in governing equation approximations. The convergence and accuracy of the MPM, including its magnetoelastic formulation, have been carefully studied and validated in previous research through both finite element analysis and experimental comparisons [67, 72, 87]. In particular, recent studies [67, 68] have experimentally validated magnetoelastic MPM simulations of quasi-static and dynamic magnetic robots under settings similar to our case study. Since the focus of the present work is on the computational optimization framework, we adopt the same modeling assumptions as in prior studies and refer interested readers to the relevant literature [67, 68] for detailed accuracy analyses.

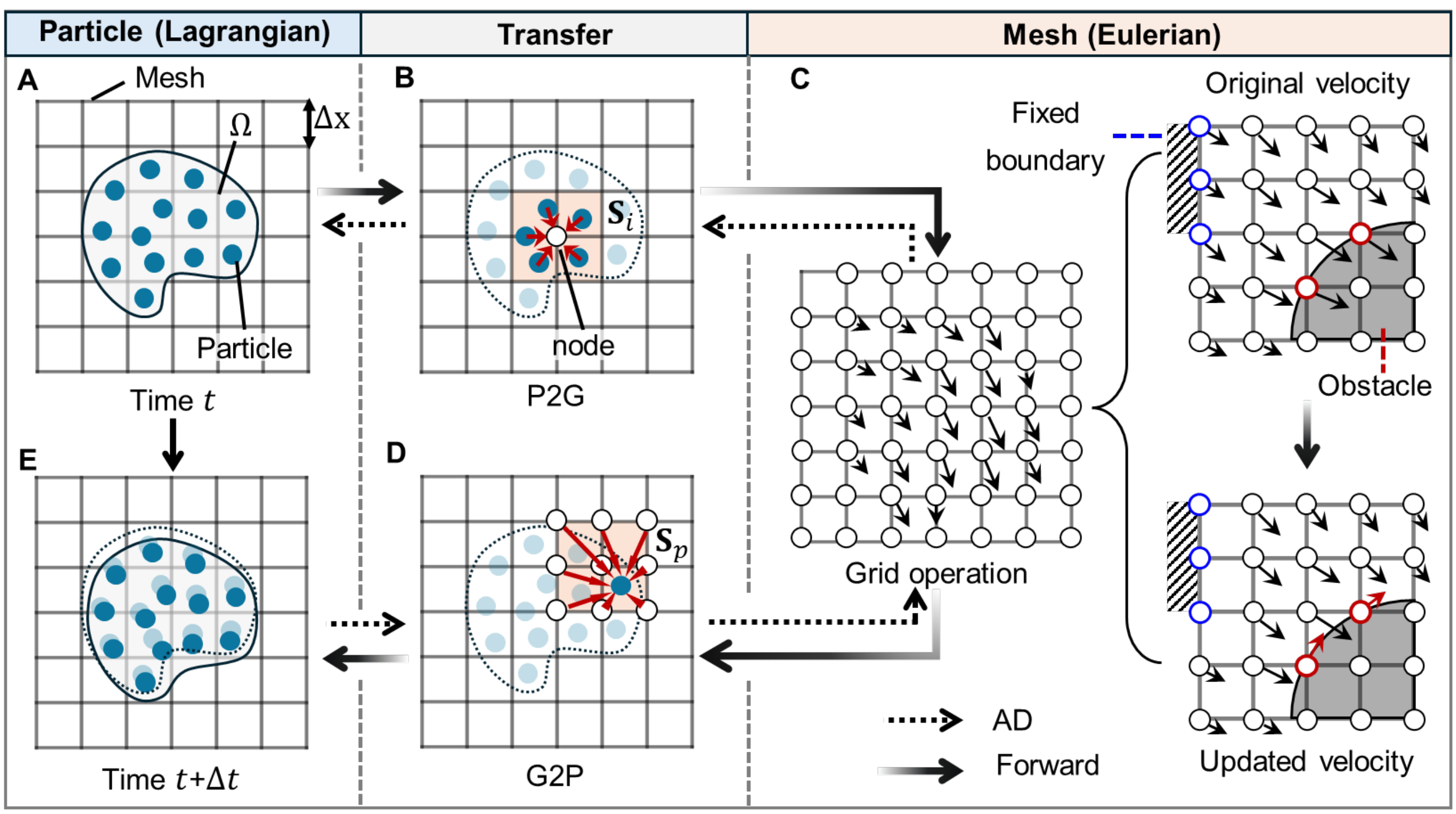

Fig. 2. Overview of the material point method. (A) The material (shaded) is represented by Lagrangian particles (blue dots) on an Eulerian background mesh. (B) Particle information is transferred to the grid node via interpolation within neighboring regions (red shades) of the node. (C) Grid operation computes

the velocity at each node (black arrows) based on the transferred particle information. The right panel shows the handling of boundaries and collisions. (D) Grid information is transferred back to the particles using the same interpolation, which is used to (E) update the particle states for the next step. The dashed arrow represents the backward propagation in the automatic differentiation (AD).

### 2.3 Parallel computing and automatic differentiation

Beyond its strengths in simulating large deformations and dynamic contacts, MPM is also particularly well-suited for parallel computing, as each iteration involves only simple, independent operations on individual particles or grid nodes. In this work, we use Taichi to implement GPU-enabled parallel computing, achieving exceptional computational performance [60]. Another key advantage of MPM is its natural support for automatic differentiation (AD). Similar to backpropagation in neural networks, AD applies the chain rule to traverse backward through the computational graph (Fig. 2), providing partial derivatives of particle states—such as position $\mathbf{x}_p(t)$ and velocity $\mathbf{v}_p(t)$ —with respect to all simulation variables like the residual magnetization ${}^p\widetilde{\mathbf{B}}_r$. This feature is highly beneficial for sensitivity analysis in topology optimization.

While sensitivity analysis can also be derived for Lagrangian FEM, it typically involves tedious and error-prone derivations. In contrast, our method automatically computes gradients, requiring only the implementation of forward simulation, with backward gradients generated almost effortlessly. Additionally, deriving analytical sensitivity becomes particularly challenging when large deformations and dynamic processes are involved, as Lagrangian meshes struggle with handling drastic morphological changes or rigid body movements. However, MPM's hybrid Lagrangian-Eulerian formulation naturally overcomes these challenges.

FEM-based sensitivity analysis also faces difficulties when dynamic contacts are present, where contact regions form, evolve, and disappear discontinuously during simulation. This makes analytical sensitivity derivation highly complicated and even intractable. In contrast, MPM effectively handles dynamic contacts using a static Eulerian mesh to facilitate gradient calculations. During grid operations, we correct grid velocities for certain contact cases by resetting their values. This adjustment breaks the chain rule by making the new grid velocity independent of the original. However, due to the inherent nature of MPM, where a particle's state variables are distributed across multiple grid nodes and then mapped back, most dependencies between variables persist across time steps even when specific grid velocities are overwritten. This enables effective sensitivity calculation, even in scenarios involving discontinuities.

## 3. Co-design framework of magnetic soft robots

In this section, we present a co-design framework for magnetic soft robots made from hard magnetic soft materials, achieving tasks such as shape morphing and locomotion. As discussed in the previous section, the actuation of magnetic soft robots involves a complex interplay between structure (material placement), material properties (magnetization), and external stimuli (applied magnetic field). For dynamic tasks, an additional dimension—time—must be considered, allowing the external magnetic field to vary throughout the process. We propose an integrated approach that simultaneously optimizes the structure, materials, and time-varying stimuli to maximize design flexibility and achieve optimal performance. First, we define a unified representation that couples all these design elements. Then, we formulate task-specific optimization problems and discuss the corresponding solutions and implementation strategies.

### 3.1 Design representation

#### 3.1.1 Structure representation

As MPM divides the structures into particle computational regions, we continue to use this particle representation in optimization. Specifically, we associate each particle $p$ with an artificial topology variable $\varphi_p \in \{0,1\}$, with zero/one representing void/solid in that particle region. To enable differentiable simulation and gradient-based optimization, we relax this binary variable to be continuous $\varphi_p \in [0,1]$ and apply a distance-based filter to smooth out the distribution shown in Fig. 3A(i-ii), addressing the checkerboard issue and controlling minimal feature size:

$$\tilde{\varphi}_p = \frac{\sum_{p' \in \mathcal{H}_p} w\big(\boldsymbol{x}_{p'}(0), \boldsymbol{x}_p(0)|R\big)\varphi_p}{\sum_{p' \in \mathcal{H}_p} w\big(\boldsymbol{x}_{p'}(0), \boldsymbol{x}_p(0)|R\big)}, \tag{24}$$

where $\mathcal{H}_p$ represents a circular or spherical neighboring region of particle $p$ within a given radius $R$ in the initial state $t = 0$, and $w\big(x_{p'}(0), x_p(0)|R\big)$ is a weight assigned to each neighboring particles

$$w\big(\boldsymbol{x}_{p'}(0), \boldsymbol{x}_p(0)|R\big) = 1 - \frac{\big|\boldsymbol{x}_{p'}(0) - \boldsymbol{x}_p(0)\big|}{R}, \tag{25}$$

which decays as the distance to the particle $p$ increases. To approximate the original binary solution with the relaxed variables, we wrap $\tilde{\varphi}_p$ with a differentiable Heaviside projection to approximate the original binary solution

$$\bar{\varphi}_p = H\big(\tilde{\varphi}_p|\beta\big) = \frac{tanh(0.5 \cdot \beta) + tanh\big[\beta \cdot \big(\tilde{\varphi}_p - 0.5\big)\big]}{2 \cdot tanh(0.5 \cdot \beta)}, \tag{26}$$

where $\bar{\varphi}_p$ is the projected topology design variable and $H(\cdot\,|\beta)$ is an approximated Heaviside function with parameter $\beta$ to control the sharpness of the projection, as shown in Fig. 3A(iii). In this study, we gradually increase $\beta$ during the optimization to drive $\tilde{\varphi}_p$ to converge to a binary value (Fig. 3A(iv)).

3.1.2 Material representation

To achieve heterogeneous responses, we spatially vary the magnetization magnitude and orientation within the materials. For each particle $p$, as shown in Fig. 3B, we define the residual magnetic flux density in the reference state ${}^p\widetilde{\mathbf{B}}_r$ to be

$${}^p\widetilde{\mathbf{B}}_r = \mathrm{B}_{r,max} \cdot {}^p\mathrm{B}_r \cdot {}^p\widetilde{\boldsymbol{b}}_r, \tag{27}$$

where $\mathrm{B}_{r,max}$ is the upper bound of the residual magnetic flux density, ${}^p\mathrm{B}_r \in [0,1]$ is the normalized magnitude design variable to control the actual residual magnetic flux density, and ${}^{p}\tilde{\boldsymbol{b}}_r = [{}^{p}\tilde{b}_{r,i} \cdot \boldsymbol{e}_i]$ is a unit vector to represent the magnetization orientation, with ${}^{p}\tilde{b}_{r,i}$ being its component along the global Cartesian axis $\boldsymbol{e}_i$. We introduce a magnetization orientation design vector $\boldsymbol{\zeta}_p = [\zeta_{p,i}] \in [-1,1]^d$, where $d$ represents the dimensionality of interest. The magnetization orientation can then be expressed as

$$ {}^{p}\tilde{b}_{r,i} = \frac{\zeta_{p,i}}{\sqrt{\sum_i \zeta_{p,i}^2 + \delta}}, \tag{28} $$

where $\delta$ is a small value to avoid singularity issues.

Although our simulation and optimization framework allow for particle-specific variations in magnetization, manufacturing such a high-resolution spatially varying magnetization distribution presents practical challenges. To accommodate the manufacturing capability, we can adopt a coarser discretization for defining the magnetization design variables ${}^{p}\mathrm{B}_r$ and $\zeta_{p,i}$. Specifically, as shown in Fig. 3B, we divide the material into $n_m$ regions $\Omega_m$ $(m = 1,2,..,n_m)$ and define the magnetization design variables for each region as ${}^{m}\mathrm{B}_r$ and $\zeta_{m,i}$. For particles within a given design region, the corresponding region's design variables are assigned:

$$ {}^{p}\mathrm{B}_r = {}^{m}\mathrm{B}_r, \qquad \zeta_{p,i} = \zeta_{m,i}, \qquad \text{for } \boldsymbol{x}_p(0) \in \Omega_m. \tag{29} $$

The particle-specific optimization can be considered a special case in which each particle serves as a separate design region, i.e., $n_m = n_p$.

3.1.3 Stimuli representation

For magnetic field excitation, we consider harmonic magnetic field stimuli (Fig. 3C), defined as:

$$ \mathbf{B}_{applied} = \mathrm{B}_{max}\mathrm{B}_a \sin\left(2\pi \bar{f} t\right) \boldsymbol{b} = \mathrm{B}_{max}\,\mathrm{B}_a \sin\left[2\pi\left((f_{max} - f_{min}) \cdot f + f_{min}\right)t\right] \boldsymbol{b}, \tag{30} $$

where $\mathrm{B}_{max}$ is the given upper bound of the applied magnetic flux density, $\mathrm{B}_a \in [0, 1]$ is a normalized magnitude design variable, $\bar{f}$ is the excitation frequency, $f_{max}$ and $f_{min}$ are given maximal and minimal frequencies, respectively, $f \in [0, 1]$ is the normalized frequency design variable, and $\boldsymbol{b} = [b_i \cdot \boldsymbol{e}_i]$ is a unit vector to represent the applied magnetic field orientation. To facilitate the orientation design, we introduce a normalized applied magnetic field orientation design vector $\boldsymbol{\eta} = [\eta_i] \in [-1,1]^d$, such that the applied field orientation is expressed as

$$b_i = \frac{\eta_i}{\sqrt{\sum_i \eta_i^2 + \delta}}, \tag{31}$$

where $\delta$ is a small numerical value to prevent division by zero.

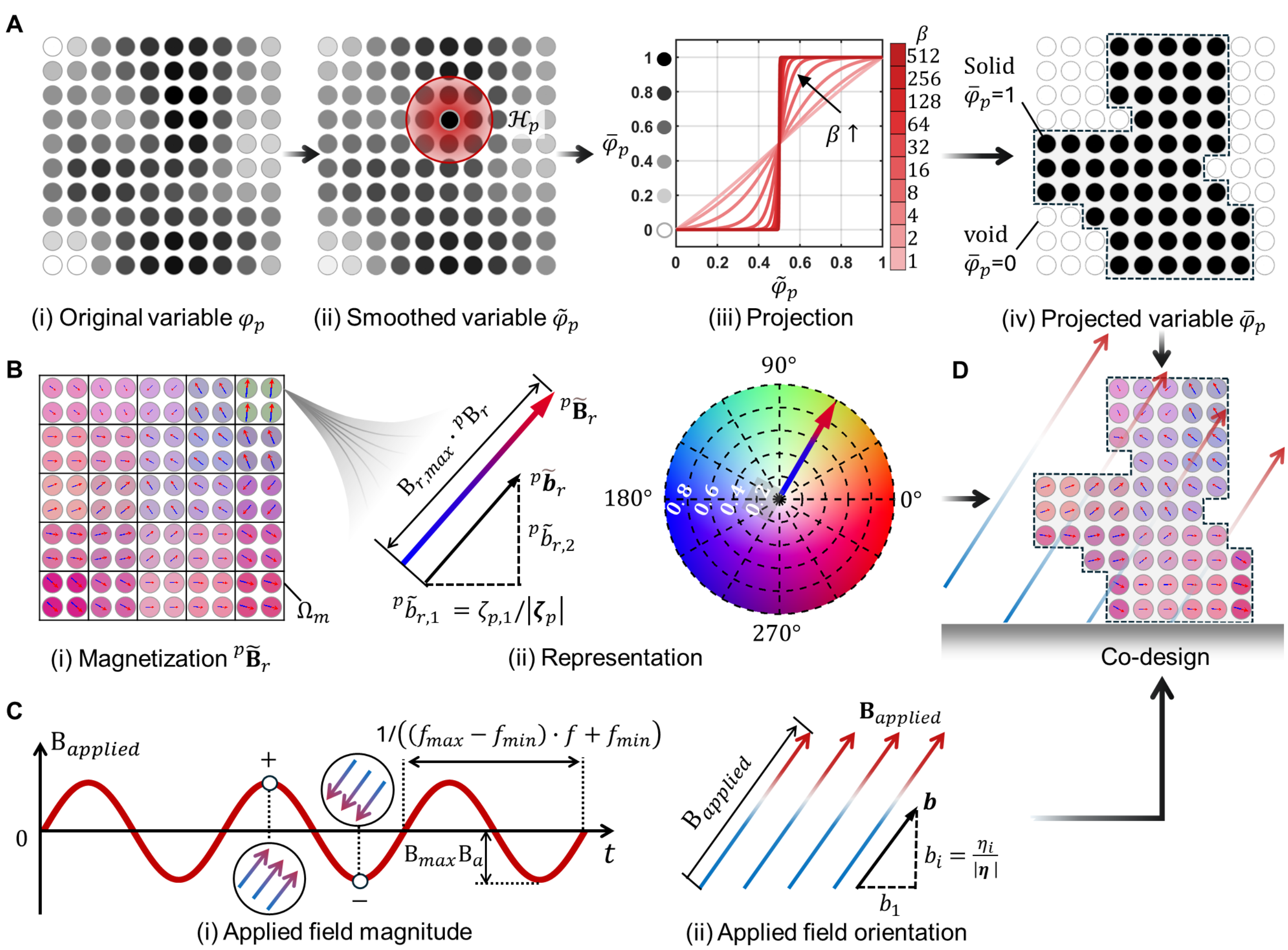

Fig. 3. Overview of the design representation. (A) Structure representation: (i) Topology is initially defined by continuous variables $\varphi_p$ at each particle. (ii) A distance-based filter smooths the topology field $\tilde{\varphi}_p$, where the red circle represents the neighboring region $\mathcal{H}_p$ and color intensity indicates weight. (iii) A Heaviside projection maps $\tilde{\varphi}_p$ to near-binary values $\bar{\varphi}_p$, with $\beta$ controlling sharpness. (iv) The final binary values define solid and void regions. (B) Material magnetization representation**:** (i) Particles

within the same design region share the same magnetization (ii) Magnetization orientation and magnitude are indicated by arrows, with length representing magnitude. These values are also color-coded at each particle for ease of visualization. (C) Stimulus representation**:** A harmonic wave encodes magnitude, while a unit vector represents orientation. Insets in the wave plot illustrate field direction changes corresponding to harmonic peaks and valleys. (D) The structure, material, and stimulus representations are integrated into the MPM simulation via the interpolated Helmholtz energy formulation, enabling the co-design process.

With design variables $\mathrm{B}_a$, $f$ and $\boldsymbol{\eta}$, we can freely control the magnitude, excitation frequency, and orientation of the applied magnetic field, enabling a time-varying magnetic stimulus. The magnetic stimulus can also take on more complex forms, such as incorporating additional frequency components or using diverse basis functions. As long as the excitation can be expressed using differentiable functions, our method remains fully applicable. However, in this study, our primary goal is to demonstrate the co-design concept for soft-magnetic robots. Therefore, for clarity and ease of illustration, we adopt this single sine-wave formulation.

3.1.4 Interpolation of the Helmholtz free energy function and particle mass

We couple all design variables in the constitutive models of hard magnetic soft materials in the MPM to simulate the actuated behaviors of a given design (Fig. 3D). This is achieved by using an interpolated Helmholtz free energy function

$$\widetilde{W}\left(\mathbf{F}|\varphi_p, {}^p\mathrm{B}_r, \zeta_{p,i}, \mathrm{B}_a, f, \eta_i\right) = \left(\epsilon + (1-\epsilon)\bar{\varphi}_p^{q_w}\right) \cdot \widetilde{W}_e(\mathbf{F}) + \bar{\varphi}_p^{q_w} \widetilde{W}_m\left(\mathbf{F}|\, {}^p\mathrm{B}_r, \zeta_{p,i}, \mathrm{B}_a, f, \eta_i\right), \tag{32}$$

where $q_w$ is the order of penalization to encourage the convergence to binary topology design variables ($q_w = 3$ in this study), $\epsilon$ is a small value to avoid singularity issues, and the magnetic potential energy can now be given as

$$\widetilde{W}_m\left(\mathbf{F}|\, {}^p\mathrm{B}_r, \zeta_{p,i}, \mathrm{B}_a, f, \eta_i\right) = -\frac{1}{\mu_0}\mathbf{F} \cdot {}^p\widetilde{\mathbf{B}}_r\left({}^p\mathrm{B}_r, \zeta_{p,i}\right) \cdot \mathbf{B}_{applied}(\mathrm{B}_a, f, \eta_i), \tag{33}$$

where ${}^{p}\widetilde{\mathbf{B}}_r({}^{p}\mathrm{B}_r, \zeta_{p,i})$ is given by Eqs. 27-29, and $\mathbf{B}_{applied}(\mathrm{B}_a, f, \eta_i)$ is given by Eqs. 30-31. The particle mass $m_p$ is given as

$$m_p = \bar{\varphi}_p \rho_0 v_p, \tag{34}$$

where $\rho_0$ is the density of the materials. With the interpolation of the Helmholtz free energy function and particle mass, we can follow the procedure detailed in Section 2 to evaluate the performance of the design and obtain gradient information via AD for later optimization.

### 3.2 Optimization problem formulations and solution procedure

The goal of our co-design framework is to simultaneously optimize structure ($\varphi_p$), materials (${}^{p}\mathrm{B}_r, \zeta_{p,i}$) and stimuli ($\mathrm{B}_a, f, \eta_i$), to improve the performance of a given task. We can formulate the optimization problem as

$$\min_{\varphi_p, {}^{p}\mathrm{B}_r, \zeta_{p,i}, \mathrm{B}_a, f, \eta_i} g(\mathbf{x}_p, \mathbf{v}_p), \tag{35a}$$

$$s.t.\ \text{Eqs. } 15-23, 32-34, \tag{35b}$$

$$0 \leq \varphi_p, {}^{p}\mathrm{B}_r, \mathrm{B}_a, f \leq 1, \tag{35c}$$

$$-1 \leq \zeta_{p,i}, \eta_i \leq 1, \tag{35d}$$

where $g(\mathbf{x}_p, \mathbf{v}_p)$ is the task-specific objective function defined on the positions and velocity of the particles, obtained from the MPM simulation summarized in Eq. 35b. In our study, we also consider multi-task optimization, in which one single design of structure and materials is used to achieve multiple tasks under different optimized stimuli. In this case, we can aggregate the objective functions together to formulate a multi-task optimization problem

$$\min_{\varphi_p, {}^{p}\mathrm{B}_r, \zeta_{p,i}, \mathrm{B}_a^{(l)}, f^{(l)}, \eta_i^{(l)}} \sum_l g^{(l)}\left(\mathbf{x}_p^{(l)}, \mathbf{v}_p^{(l)}\right), \tag{36a}$$

$$s.t.\ \text{Eqs. } 15-23, 32-34, \tag{36b}$$

$$0 \leq \varphi_p, {}^{p}\mathrm{B}_r, \mathrm{B}_a^{(l)}, f^{(l)} \leq 1, \tag{36c}$$

$$-1 \leq \zeta_{p,i}, \eta_i^{(l)} \leq 1, \tag{36d}$$

where $l$ represents the task label and is used as a superscript to denote variables corresponding to that specific task. These optimization problems are solved iteratively, as outlined in Algorithm 1.

**Algorithm 1** Solution procedure for the co-design problem in Eq. 36

**Input:**

Initial values: $\varphi_p$ (structural design variables)
${}^p\mathrm{B}_r, \zeta_{p,i}$ (materials design variables)
$\mathrm{B}_a^{(l)}, f^{(l)}, \eta_i^{(l)}$ (stimuli design variables)
$R$ (filter radius in Eq. 25)
$\beta$ (projection parameter in Eq. 26)
$\beta_{max}$ (maximal projection parameter)
$N_\beta$ (interval for projection update)
$q_w$ (penalty coefficient in Eq. 32)
$N_{max}$ (maximal iterations)

**Output:**

Optimized design variables $\boldsymbol{\Theta} = \{\varphi_p, {}^p\mathrm{B}_r, \zeta_{p,i}, \mathrm{B}_a^{(l)}, f^{(l)}, \eta_i^{(l)}\}$

Initialize iteration counter $n_{iter} \leftarrow 0$ ▷ Initialize
**while** $\boldsymbol{\Theta}$ not converged and $n_{iter} < N_{max}$ **do**
  $n_{iter} \leftarrow n_{iter} + 1$ ▷ Update iteration counter
  $g \leftarrow 0$, $\partial g/\partial \boldsymbol{\Theta} \leftarrow 0$ ▷ Reset objective and sensitivity
  Obtain filtered $\tilde{\varphi}_p$ via Eq. 24 ▷ Distance-based filter
  $\bar{\varphi}_p \leftarrow H(\tilde{\varphi}_p \,|\, \beta)$ via Eq. 26 ▷ Heaviside projection
  **for** $l = 1$ to $N_l$ **do** ▷ For each of $N_l$ tasks
    Calculate $\mathbf{x}_p^{(l)}, \mathbf{v}_p^{(l)}$ via Eqs. 15-23,32-34 ▷ Forward simulation
    Object for $l^{th}$ task $g^{(l)}(\mathbf{x}_p^{(l)}, \mathbf{v}_p^{(l)})$ ▷ Design evaluation
    Calculate sensitivity $\partial g^{(l)}/\partial \boldsymbol{\Theta}$ ▷ Backward AD
    $g \leftarrow g + g^{(l)}$ ▷ Accumulate the objective
    $\partial g/\partial \boldsymbol{\Theta} \leftarrow \partial g/\partial \boldsymbol{\Theta} + \partial g^{(l)}/\partial \boldsymbol{\Theta}$ ▷ Accumulate the sensitivity
  **end for**
  $\boldsymbol{\Theta} \leftarrow Adam(\boldsymbol{\Theta}, \partial g/\partial \boldsymbol{\Theta})$ ▷ Adam solver to update $\boldsymbol{\Theta}$
  **if** $\beta < \beta_{max}$ and $n_{\mathrm{iter}}$ mod $N_\beta = 0$ **then** ▷ For every $N_\beta$ iterations
    $\beta \leftarrow 2\beta$ ▷ Increase projection sharpness
  **end if**
  Evaluate the convergence of $\boldsymbol{\Theta}$
**end while**

In each iteration, the full motion history of the current robot design, including position and velocity at each timestep, is computed via forward simulation using the magneto-elastic MPM. This motion history is then used to evaluate the aggregated objective function, which quantifies the robot's performance.

Sensitivities of the objective function with respect to the design variables are obtained through automatic differentiation within the MPM framework. Based on these sensitivities, the design variables are updated within their prescribed bounds. Since the optimization problem does not involve large-scale nonlinear constraints, we can directly use the gradient-based optimizer Adam for this iterative update to fully leverage GPU parallelization. The updated design is passed to the next iteration until convergence or until the maximum number of iterations, $N_{max}$, is reached. Every $N_{\beta}$ iterations, we increase the projection parameter $\beta$ to gradually sharpen the binary projection of the topology variable, until it reaches a given upper bound $\beta_{max}$. To assess convergence, we compute the average objective function value over the most recent 10 iterations and compare it to the average from the preceding 10 iterations. The relative absolute change between these averages is calculated, and the optimization is considered converged if this relative difference is smaller than 0.001. In this study, the simulation and optimization are implemented on a single NVIDIA RTX 6000 Ada GPU (equipped with 48GB GDDR6 graphics memory and 18,176 CUDA cores).

# 4. Design case study

## 4.1 Quasi-static designs for hard magnetic soft continuum robots

In this case study, we focus on the quasi-static design of magnetic soft continuum robots (MSCRs) [7, 17, 18, 32, 33]. An MSCR comprises a non-magnetized body that advances or retracts via a motor and a magnetically active distal portion, which is essentially a slender rod made of hard-magnetic soft materials that bends in response to magnetic fields (Fig.4). The active bending of the distal portion makes MSCRs highly promising for navigating complex vasculature through remote magnetic control in surgical applications. Here, we apply the proposed framework to design the magneto-active distal portion, targeting two key tasks for MSCR operation: maximizing deformation and achieving shape morphing (Fig. 4).

Since the distal portion primarily operates in a constant magnetic field, we neglect harmonic terms in the stimuli and focus on its magnitude and orientation for quasi-static behaviors. Given that the MSCR is highly slender, with a length of 17.5 mm and a height of 0.85 mm, topology design is unnecessary in this case. Instead, we concentrate on co-designing the applied magnetic field and the remanent magnetization of the materials. The robot is horizontally divided into 10 equal-sized design regions $\Omega_m$, with all particles within each region sharing the same magnetization. The materials have an average shear modulus of $G = 303\ kPa$ and a maximum remanence of $\mathrm{B}_{r,\,max} = 0.143$ T. The bulk modulus is set to $K = 1000G$ to approximate near incompressibility. For the MPM simulation, we generate particles on a uniform grid within the robot. We set the settling time $\mathrm{T}$ of the simulation to 0.05s with a damping coefficient of $c = 200\ \mathrm{Ns/m}$ to stabilize the system and accelerate convergence. These settings have been sufficient to ensure convergence to a stable static state in all cases.

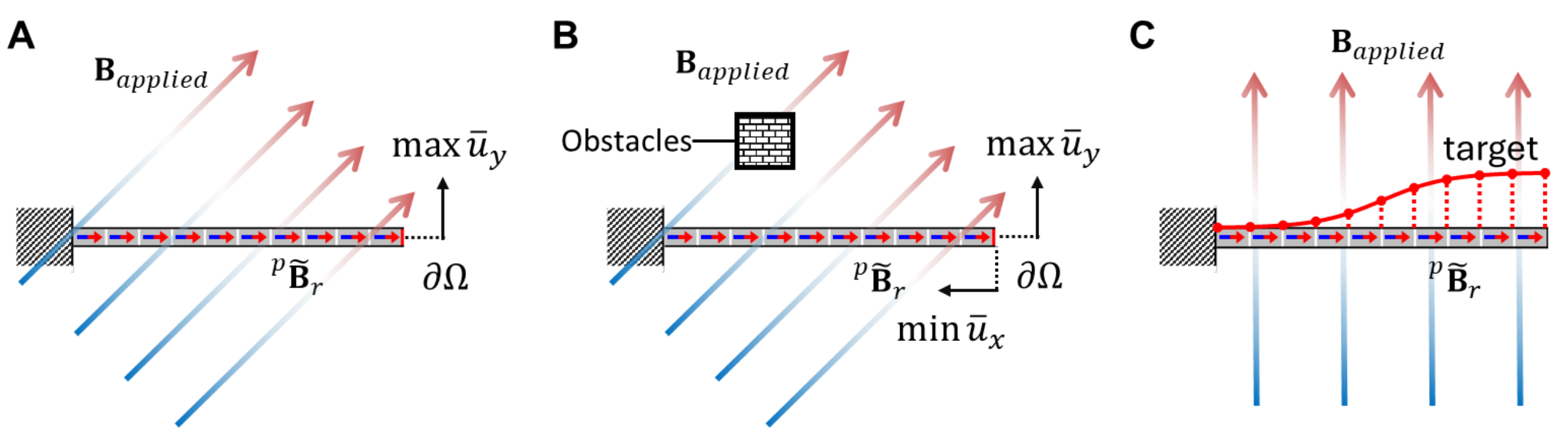

Fig. 4. Quasi-static design cases. (A) Maximizing the tip height without obstacles, where the shaded region represents the fixed boundary. (B) Maximizing the tip height or shifting the tip as far left as possible in the presence of obstacles. (C) Shape morphing task, with the red curve indicating the target shape.

4.1.1 Single objective design

We applied our framework to a single-objective design problem, as shown in Fig. 4A and Fig. 5. The maximal magnitude of the applied magnetic field is $\mathrm{B}_{a,max} = 8$ mT. In this case, the goal is to maximize the height reached by the region of interest $\partial\Omega$. The optimization objective can be formulated as:

$$\min_{{}^{m}\mathrm{B}_r, \zeta_{m,i}, \mathrm{B}_a, \eta_i} g = -\bar{u}_y = -\frac{1}{n_{\partial\Omega}} \sum_{p\in\partial\Omega} \left(\mathrm{x}_{p,2}(\mathrm{T}) - \mathrm{x}_{p,2}(0)\right), \tag{37}$$

where $\bar{u}_y$ is the average displacement of the tip along the y-axis, $n_{\partial\Omega}$ is the number of particles on $\partial\Omega$, and $\mathrm{x}_{p,2}(\mathrm{T})$ represents the y-coordinates of these particles at time T. Note that the original maximization problem is transformed to a minimization in the formulas by switching the sign of $\bar{u}_y$. The initial design (Fig. 5) has an applied magnetic field at a 45-degree angle to the horizontal axis with $\mathrm{B}_a = \mathrm{B}_{a,max}$, while all particles have their remanent magnetization aligned horizontally with magnitude ${}^{m}\mathrm{B}_r = \mathrm{B}_{r,max}$. The embedded magnetic particles tend to align with the external field, generating a distributed magnetic torque that deforms the robot. In the resulting stable state, the free end of the robot aligns closely with the external field. At this point, the magnetic torque is insufficient to further overcome elastic resistance and increase the tip height. This actuation process involves large deformation, which is challenging to simulate by Lagrangian FEM, but it can be easily handled by our method and finished within three seconds.

When optimizing only the external stimulus (i.e., orientation $\eta_i$ and magnitude $\mathrm{B}_a$), the magnetic field rotates and points to the left. In the stable state, a significant portion of the particles maintain a magnetization almost perpendicular to the external field, generating a strong magnetic torque to counteract elastic resistance and keep the free end of the robot nearly vertical, maximizing its height. If we optimize only the material properties (i.e., orientation $\zeta_{m,i}$ and magnitude ${}^{m}\mathrm{B}_r$), we observe that in the optimized design, the right half has its magnetization oriented opposite to the applied magnetic field in the unactuated reference state, while the left half is perpendicular to the field. As the actuation progresses, the left half contributes the most in the initial stages. When the free end reaches a vertical state, the embedded magnetic particles in the right half also become perpendicular to the external field, generating a large magnetic torque that counteracts the elastic deformation. Finally, when co-designing both materials and stimuli, the optimized magnetic field points upward, while the magnetization orientation transitions gradually from left to right. This produces a spatially varying torque that maintains a balance with the elastic resistant forces, ultimately resulting in a similar vertical free end. In this case study, all design approaches achieve nearly the same maximum height.

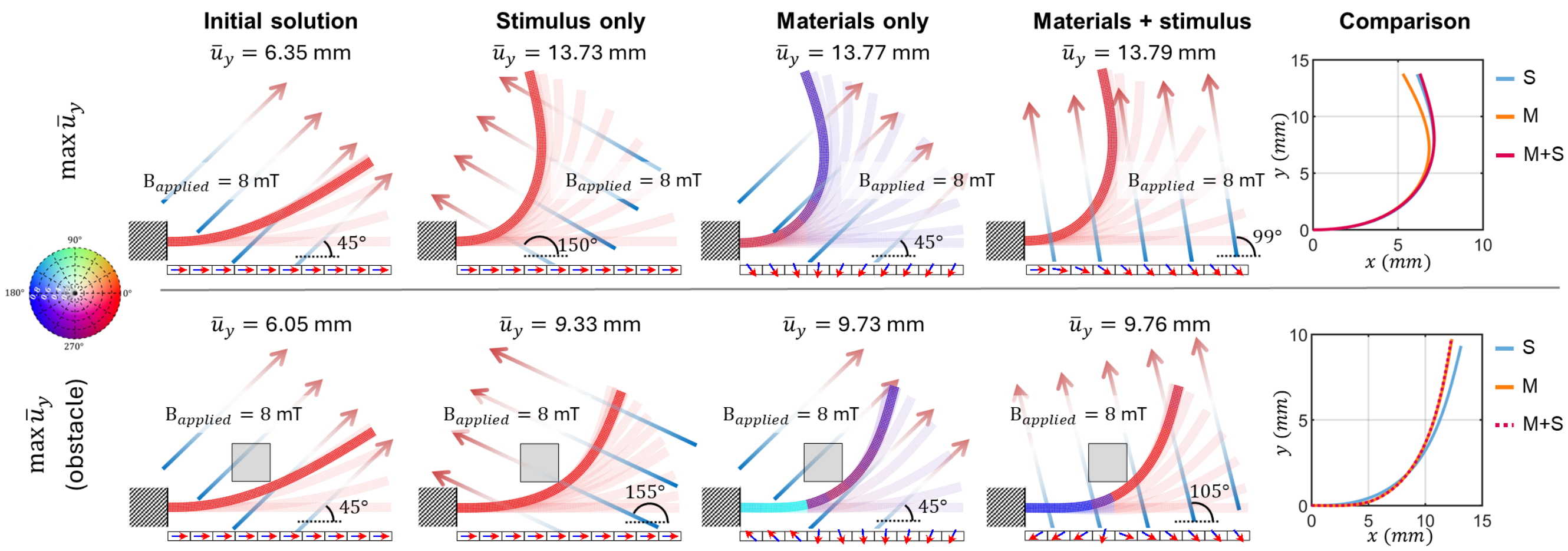

Fig. 5. Design results for maximizing the tip height. The first and second rows correspond to cases without and with obstacles, respectively. The first four columns show: (i) the initial design before optimization, (ii) the design optimized only for stimulus, (iii) the design optimized only for materials, and (iv) the co-design of both materials and stimuli. In each figure, the solid body represents the final state, while transparent shades depict intermediate states. The last column directly compares the robots' centerlines. Particle colors indicate magnetization orientation and magnitude, as shown in the small polar plot inset. To enhance visualization, each figure includes a bottom inset with arrows representing magnetization orientation and magnitude (proportional to arrow length).

We now explore the second case (Fig. 4B), where all settings remain the same except for the addition of an obstacle. In this scenario, the actuated initial design collides with the obstacle during upward bending, preventing it from reaching greater heights (second row in Fig. 5). Capturing such dynamic contact is challenging with Lagrangian FEM and may complicate sensitivity analysis. In contrast, our method successfully captures contact and provides gradient information, allowing us to continue using gradient-based optimization. Specifically, when optimizing only the stimulus (Fig. 5), the applied magnetic field still points left but is closer to horizontal. This alignment maximizes the magnetic torque at the free end, encouraging bending and allowing the robot to reach higher than the initial design. When optimizing only the materials, the magnetization is adjusted so that nearly all segments have magnetization aligned perpendicular to the external field, generating maximum torque. Notably, segments near the fixed end exhibit a flipped magnetization orientation compared to the obstacle-free case. This causes downward bending near the fixed end, increasing curvature to compensate for the obstacle's resistance. A similar effect is observed in the co-design of materials and stimulus (Fig. 5). As a result, material-only design and co-design outperform stimulus-only design, highlighting that stimulus alone lacks the ability to achieve spatially varying deformation control. In contrast, modifying material distribution better accommodates local constraints, which becomes crucial when obstacles or other constraints are present.

The optimization histories for the above cases are shown in Fig. 6. In all scenarios, both with and without obstacle, the optimization converges rapidly, with a significant objective drop occurring within the first few iterations. When designing only the materials or only the stimuli, the convergence is relatively smooth. In contrast, the co-design of materials and stimuli shows small fluctuations due to the increased complexity of the design space and the materials-stimuli interplay. Nevertheless, the optimization algorithm reliably reaches convergence in all cases. For most cases, the optimization converged within 30 to 40 iterations, with each iteration taking approximately 7.7 seconds. As a result, the entire optimization process was completed within five minutes. This is exceptionally efficient, given the nonlinear physics, dynamic contacts, and the large design space involved.

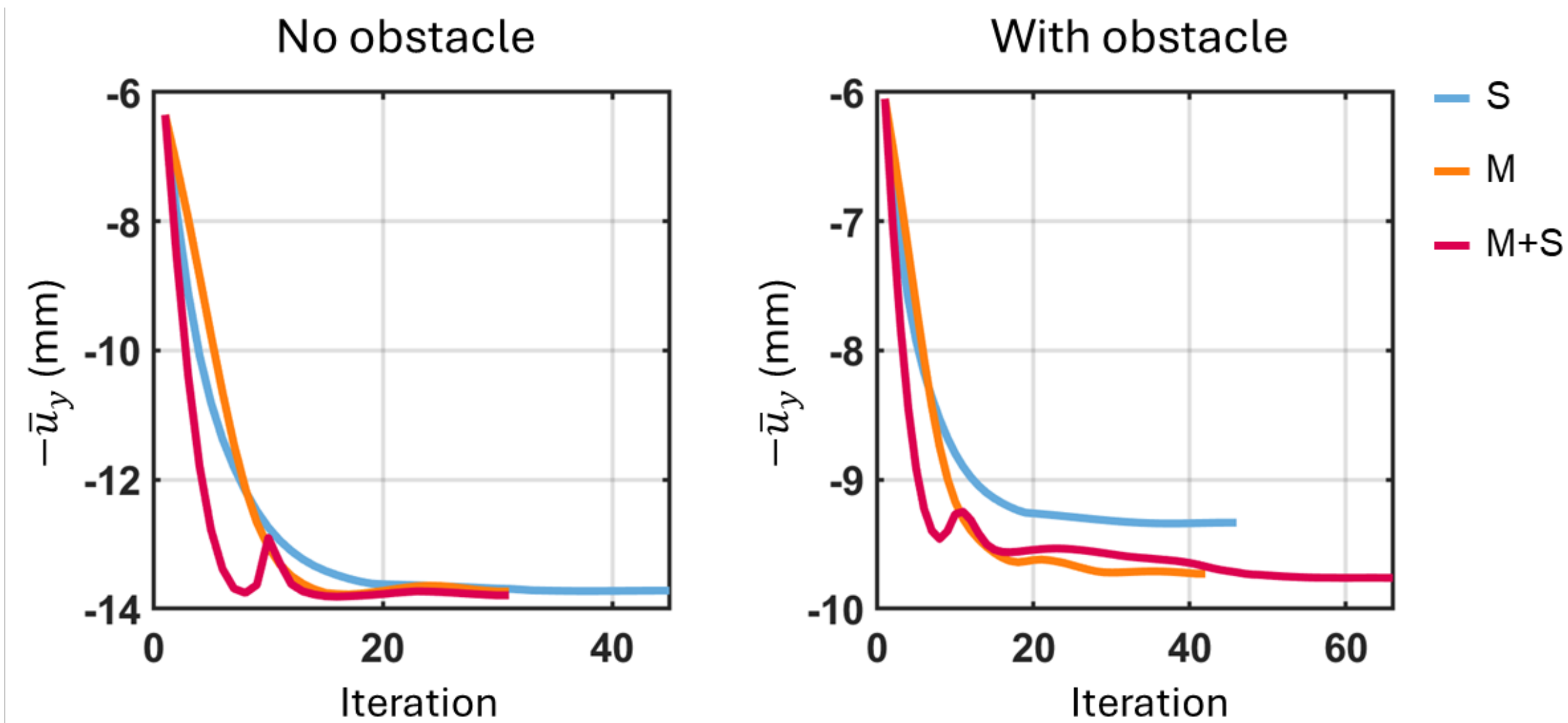

Fig. 6. Optimization histories for the cases aiming to maximize tip height in Fig. 5.

To further illustrate the difference between stimulus-only and material-only designs, we consider two additional cases (Fig. 4B-C). For the case in Fig. 4B, we change the maximal magnitude of the applied magnetic field to $\mathrm{B}_{a,max} = 28$ mT, keep other settings the same, but aim to move the tip as far left as possible, with the optimization function given as:

$$\min_{{}^{m}\mathrm{B}_r,\zeta_{m,i},\mathrm{B}_a,\eta_i} g = \bar{u}_x = \frac{1}{n_{\partial\Omega}} \sum\nolimits_{p\in\partial\Omega} \left(\mathrm{x}_{p,1}(\mathrm{T}) - \mathrm{x}_{p,1}(0)\right). \tag{38}$$

For the case in Fig. 4C, there are no obstacles, and the objective is to achieve a shape morphing that matches a given target shape. The maximal magnitude of the applied magnetic field is changed to $\mathrm{B}_{a,max} = 10$ mT. The optimization function is defined as:

$$\min_{{}^{m}\mathrm{B}_r,\zeta_{m,i},\mathrm{B}_a,\eta_i} g = \mathrm{MSE}(\mathbf{x}_p(\mathrm{T}), \bar{\mathbf{x}}_p)\,, \text{for } p \in \partial\Omega, \tag{39}$$

where MSE is the mean square error, $\bar{\mathbf{x}}_p$ is the shape target, and $\partial\Omega$ is the center line. In both cases, achieving optimal performance requires spatially varying local deformation, which demands greater material heterogeneity, something global stimulus control alone cannot provide. The design results are shown in Fig. 7.

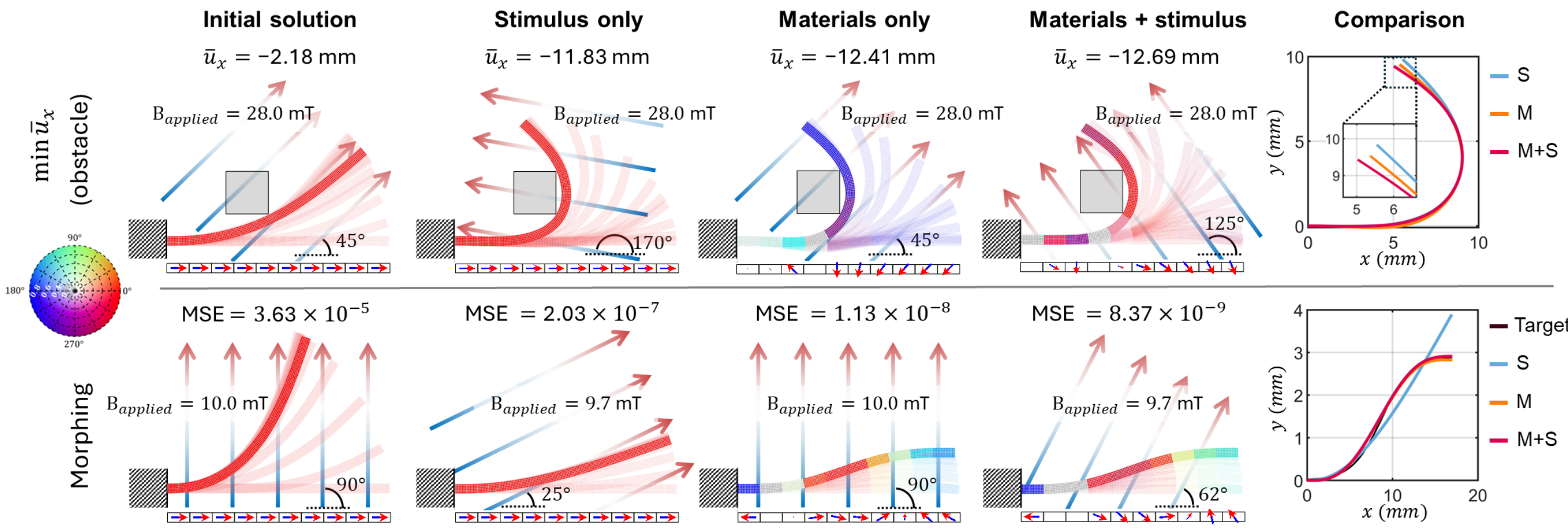

Fig. 7. Design results for the left reaching (first row) and morphing (second row) tasks.

Specifically, for the left-reaching task, the optimized robots are divided into three segments by the two contact points, each segment experiencing drastically different boundary conditions. Therefore, distinct torques are required in these segments to achieve a large curvature for bypassing the obstacle. A stimulus-only design lacks the ability to accommodate such heterogeneous local requirements, resulting in poor performance. In contrast, both the material-only and co-design approaches introduce the necessary spatial heterogeneity within the robot, with weak magnetization at the fixed end gradually transitioning to stronger magnetization at the free end, where the orientation is nearly perpendicular to the applied magnetic field in the actuated state. This enables greater curvature around obstacles, outperforming the stimulus-only design. Notably, the co-design approach achieves the best performance, as it offers the highest flexibility among the three designs. Similarly, in the shape-morphing task, the robot requires counterclockwise bending at the fixed end and clockwise bending at the free end. Stimulus-only design struggles to achieve this, while material-only and material-stimulus co-design approaches provide the necessary flexibility, resulting in significantly better performance.

4.1.2 Multi-objective design

To further demonstrate the importance of co-design, we explore a multi-objective design in which a single robot must perform two distinct tasks under different external magnetic fields. Specifically, we aim to maximize the tip height (Fig. 4B) and achieve the shape-morphing task (Fig. 4C) under two different excitations. While the applied magnetic fields are optimized separately for each task, the material distribution is shared. The optimization is formulated by aggregating the two objectives using the equation in Eq. 36. This multi-objective task is significantly more complex than in previous cases, as the two objectives inherently compete. For the tip height maximization, the robot requires a strong counterclockwise bend throughout its structure to bypass the obstacle. In contrast, the shape-morphing task demands a gradual transition from counterclockwise to clockwise bending, progressing from left to right.

Both stimulus-only and material-only designs fail to achieve a good trade-off between these objectives. The stimulus-only design excels in the height maximization task but performs poorly in shape morphing. Meanwhile, the material-only design struggles to balance the two tasks, leading to suboptimal performance in both. Notably, in height maximization, the material-only design behaves similarly to the morphing task, meaning it sacrifices height maximization to prioritize shape morphing, which better aligns with its strengths in achieving local heterogeneity.

In contrast, the co-design approach successfully achieves strong performance in both tasks. By leveraging a heterogeneous material distribution, it enables spatially varying bending for the morphing task. For height maximization, the applied magnetic field is adjusted to transform the originally clockwise-bending free end in the morphing task into counterclockwise bending. This co-design achieves the best balance between both objectives. It highlights the critical role of co-design in enhancing flexibility for multi-functionality, which is made possible by our proposed framework.

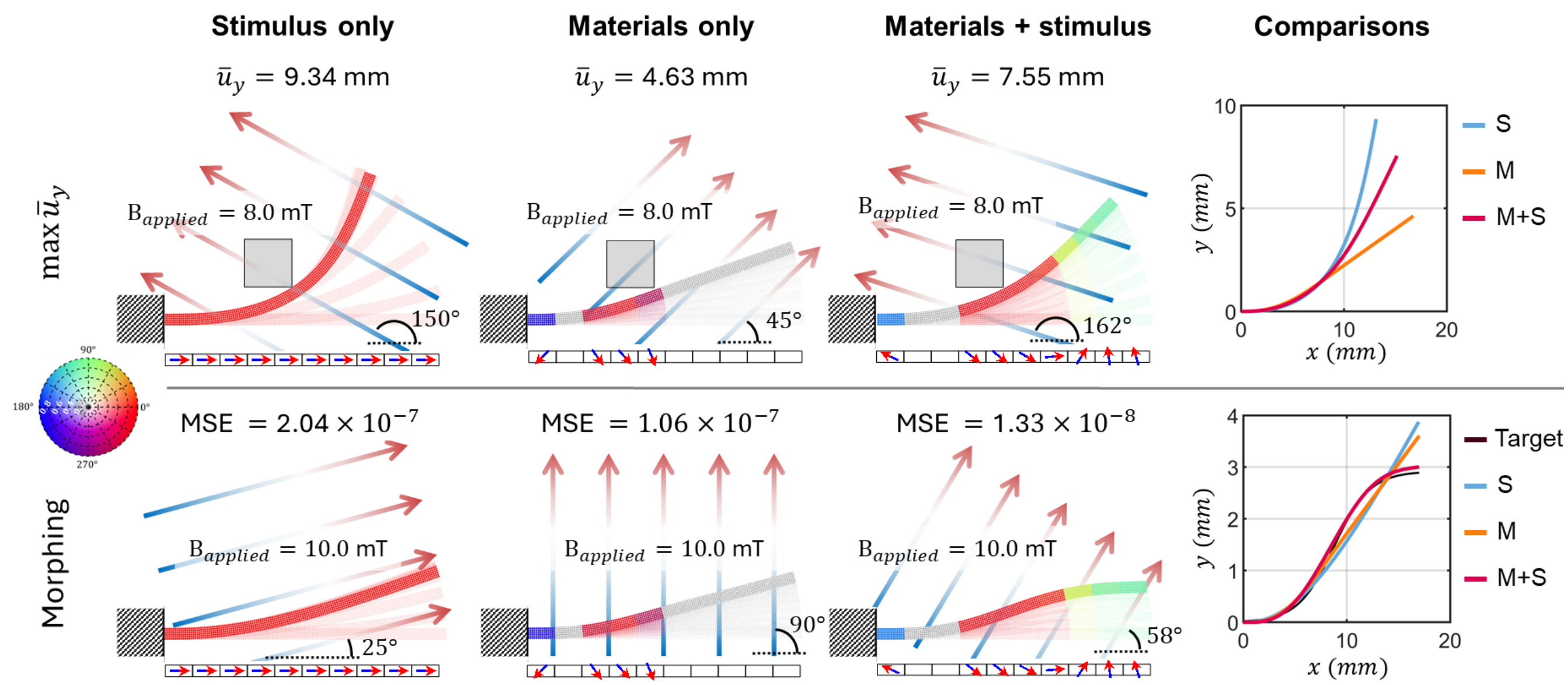

Fig. 8. Multi-task design results. The two rows correspond to the two different tasks. While the magnetic field varies, robot designs are shared between the tasks.

## 4.2 Dynamic designs for magnetic soft robot locomotion

Beyond handling large deformations and contact interactions, our method also accommodates dynamic applications that involve rigid body motions, time-varying stimuli, and changing boundary conditions. In this subsection, we explore the dynamic cases by designing various magnetic soft robot locomotion, including crawlers and walkers (Fig. 8). Both tasks aim to maximize locomotion distance within a given timeframe, but due to differences in aspect ratios, they are expected to exhibit distinct movement patterns. The slender design in Fig. 8A is likely to adopt a crawling motion, while the thicker structure in Fig. 8B is expected to rely on a walking motion, hence the terms "crawler" and "walker." For the crawler, we simultaneously optimize the harmonic applied magnetic field (including frequency, magnitude, and orientation) and the material properties (including the magnitude and orientation of remanent magnetization). For walkers, in addition to stimulus and material design, we also co-design the structure using the topology variables. All cases consider gravity, with the hard magnetic soft materials having an average shear modulus of $G = 303$ kPa and a maximum remanence of $\mathrm{B}_{r,\,max} = 0.05$ T. The damping coefficient is set to $c = 200$ Ns/m, and the bulk modulus is set to $\mathrm{K} = 1000G$ to approximate near incompressibility.

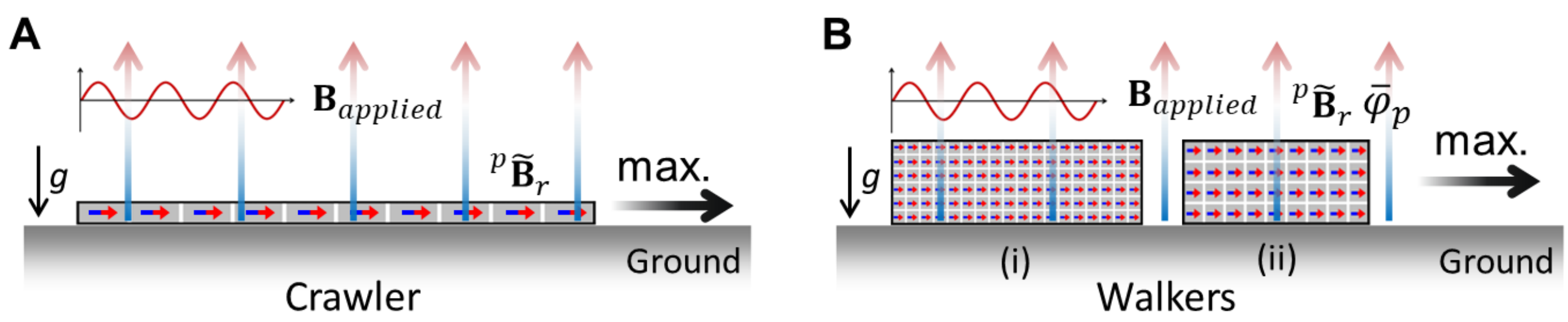

Fig. 9. Magnetic soft robot locomotion design cases. (A) Crawler: Co-designing the applied magnetic field and material magnetization within the grid-like design regions. (B) Walkers: Co-designing the applied magnetic field, material magnetization within the grid-like design regions, and particle-specific topology variables. (i) and (ii) represent two walkers with different aspect ratios.

4.2.1 Materials-stimulus co-design for crawlers

For the crawler design, we consider a slender structure with a length of 10 mm and a height of 0.5 mm. As with the previous MSCR cases, topology design is unnecessary here due to the slenderness, so we focus on co-designing the applied magnetic field and the remanent magnetization of the materials. The robot is horizontally divided into 10 equal-sized design regions $\Omega_m$, with all particles within each region sharing the same magnetization. We assume a harmonic excitation defined in Eq. 30, with the $\mathrm{B}_{max} =$ 5 mT, $f_{min} = 10$ Hz , and $f_{max} = 100$ Hz. Since the excitation is periodic, we only need to consider a time range that covers multiple cycles. Here, we assume $\mathrm{T} = 0.2\ s$, and time step $\Delta t = 2 \times 10^{-6} s$. The optimization problem is formulated as:

$$\min_{{}^{m}\mathrm{B}_r, \zeta_{m,i}, \mathrm{B}_a, f, \eta_i} g = -\sum_{t=0}^{T} \bar{\mathbf{v}}(t) \cdot \boldsymbol{e}_1 \Delta t\,, \tag{40}$$

where $\bar{\mathbf{v}}(t)$ is the average velocity of the whole crawler. We evaluate the design under different surface roughness conditions: smooth ($\mu = 0.25$), medium ($\mu = 0.5$), and very rough (non-slip), with results shown in Fig. 9. Although optimization is autonomous without any human guidance, the results reveal an interesting pattern—all crawlers evolve a locomotion mode characterized by a traveling wave propagating from tail (left end) to head (right end). This results in the characteristic "hump" motion commonly observed in caterpillars [88].

On smooth surfaces, the crawling wave has a larger magnitude and wavelength, as shown in Fig. 9A. It starts with the tail lifting due to a clockwise magnetic torque, and, as the magnetic field reverses direction, the tail quickly curves downward, arching the whole body up to build momentum and form a hump in the middle. With the tail anchoring the body, the wave propagates forward, allowing the body to be propelled. Once the hump reaches the head and the whole body relaxes, another cycle begins with the magnetic field switching once again. Through this cyclic motion, the wave converts time-varying torque into forward momentum, enabling the crawler to move forward. Interestingly, this resembles the backward crawling mode of caterpillars, where most of the body is airborne for efficient movement [88].

On rough surfaces (Fig. 9C), the crawling wave exhibits a smaller magnitude and wavelength, corresponding to a higher excitation frequency. Rather than arching the entire body off the ground, each segment is lifted sequentially, moved forward into the adjacent segment, and then lowered back onto the surface. This motion resembles the forward crawling of a caterpillar, where most legs remain in contact with the ground, enhancing friction and stability at the cost of momentum [88]. For medium friction, the behavior falls between these two extremes, as shown in Fig. 9B.

Throughout wave propagation in all cases, different regions of the robot make contact with the ground, constantly shifting the contact regions. Under time-varying external excitation, this results in a complex interaction between the robot, stimuli, and environment. Our method effectively manages these dynamics by generating heterogeneous material distributions and optimizing the full locomotion cycle to adapt to varying surface conditions, demonstrating exceptional flexibility. For all crawler designs, convergence was achieved in 68 to 142 iterations, with each iteration taking approximately 15 seconds. Consequently, the entire optimization process for this complex dynamic system was completed within 17 to 37 minutes.

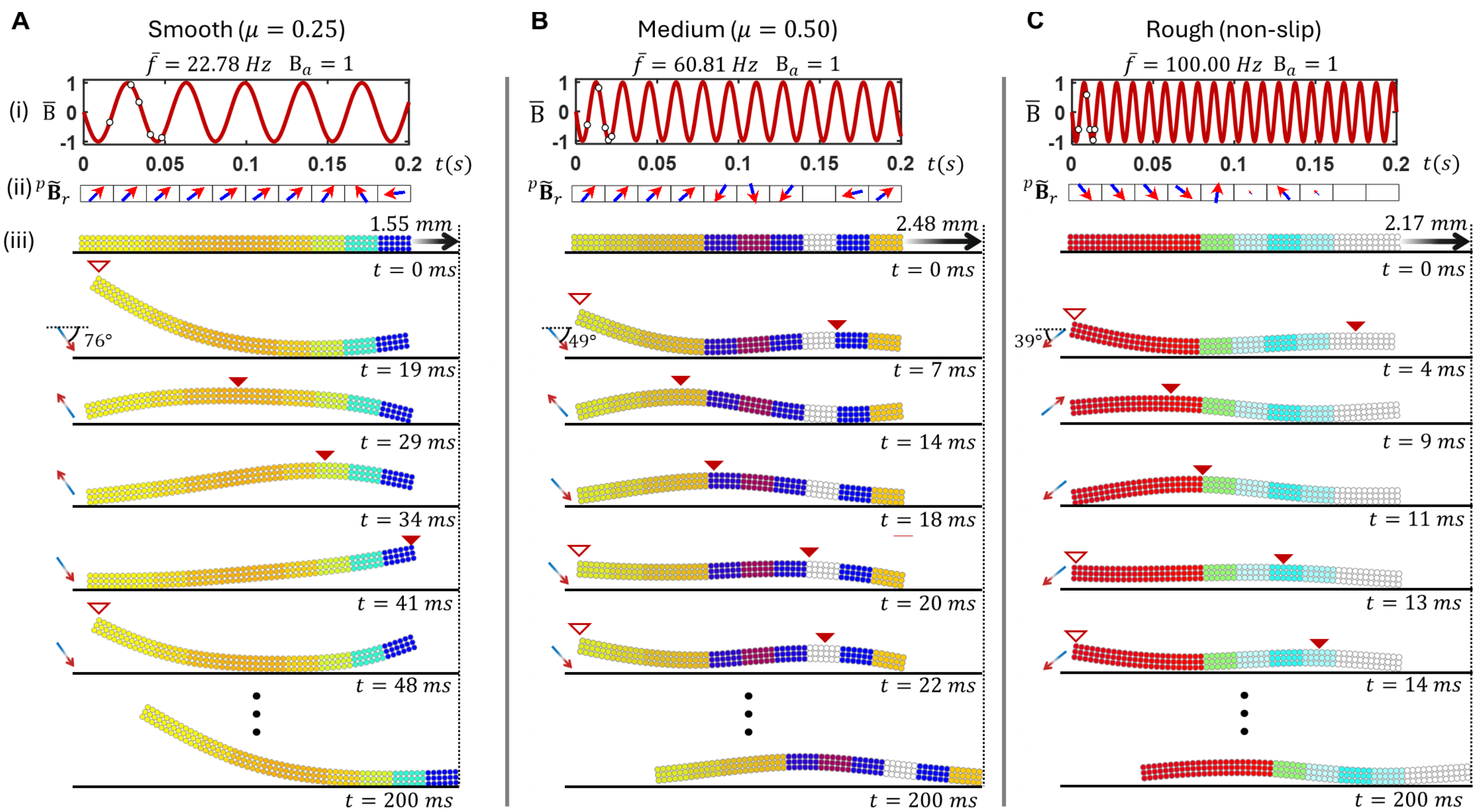

Fig. 10. Design results for the crawlers on different surfaces. (A) Smooth surface. (B) Surface with medium roughness. (C) Very rough surface with a non-slip condition. For each case, from top to bottom: (i) The optimized harmonic magnitude, shown as the red solid curve. The vertical axis is the normalized magnitude $\overline{\mathrm{B}} = \mathrm{B}_{applied}/\mathrm{B}_{max}$. (ii) Optimized magnetization. (iii) Snapshots of the locomotion history with the same color scheme as in Fig. 3B(ii). In addition to the starting and final positions, five intermediate states are selected to illustrate the behaviors in a cycle, corresponding to the five dots in the harmonic curve plot. The orientation of the applied magnetic field in each state is marked in the lower left corner. The moving distance is indicated at the starting state. For each intermediate state, the orientation of the applied field is shown. Void triangles mark the formation of the hump, while solid triangles indicate

the position of the formed hump.

4.2.2 Structure-materials-stimulus co-design for walkers

For the walker design, we consider two different body shapes, as shown in Fig. 8B: one with a height-to-length ratio of 1:3 and the other with 1: 2.25, both with a height of 2 mm. These are divided into $6 \times 18$ and $4 \times 9$ equal-size regions for material magnetization, respectively. Unlike the crawler case, we also optimize the structure by introducing a topology variable for each particle. While magnetization is designed for each element within the grid-like design regions to ensure practical fabrication, the topology variable can vary for each particle. The optimization problem is modified to incorporate these particle-specific topology variables $\varphi_p$:

$$\min_{m_{\mathrm{B}_r}, \zeta_{m,i}, \mathrm{B}_a, \eta_{i}, f, \varphi_p} g = -\sum_{k=0}^{T/\Delta t} \bar{\mathbf{v}}(k\Delta t) \cdot \boldsymbol{e}_1 \Delta t\,, \tag{41}$$

Since topology changes during the iterations, we use the mass-averaged velocity $\bar{\mathbf{v}}(t)$ to avoid any bias toward the total mass. Every 10 iterations, we update the β parameter to gradually sharpen the binary projection of the topology variable, following the scheme illustrated in Fig. 3A(iii), until it reaches 128. All other settings remain the same. The design results are shown in Figs. 11-14.

In the first case, where the structure has a larger length-to-height ratio (Fig. 10), topology optimization results in two limb-like structures (Fig. 11). The back limb is significantly larger and exhibits a stronger magnetization, primarily oriented in a horizontal direction. This orientation is nearly perpendicular to the optimized vertical magnetic field, generating a large magnetic torque. In contrast, the forelimb is thinner, has weaker magnetization, and is mostly aligned with the external magnetic field, resulting in a much smaller magnetic torque.

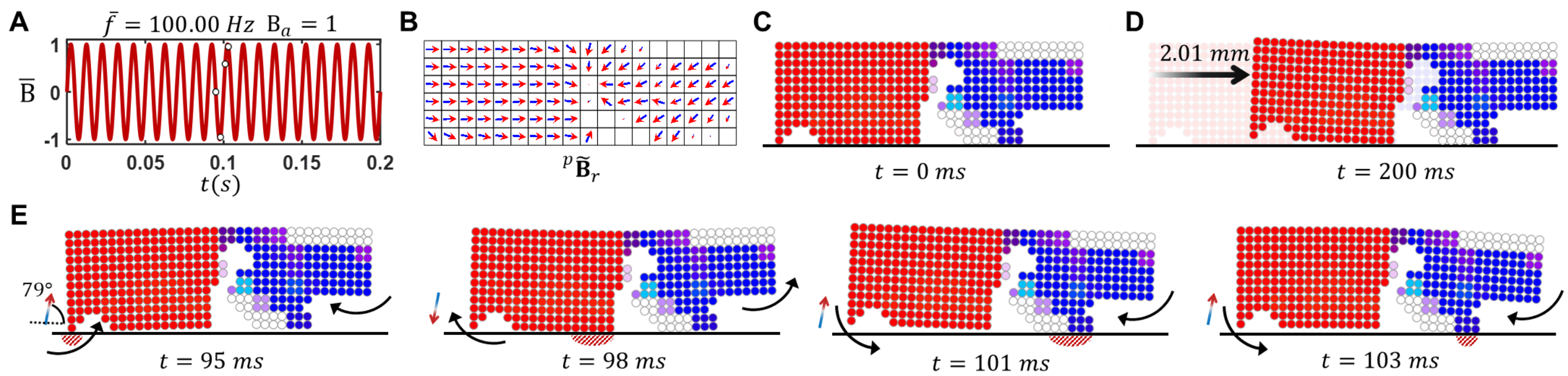

Fig. 11. Design results for the first walker. (A) Normalized magnitude of the optimized applied magnetic field. (B) Optimized magnetization. (C) Starting state, with the same color scheme as in Fig. 3B(ii). (D) Ending state, where the transparent shape shows the starting state. The arrow marks the total moving distance. (E) Snapshots of a locomotion cycle, corresponding to the dots in Fig. 11A. The orientation of the applied magnetic field is shown in the lower left, with the black arrow indicating the bending direction. The patterned shapes represent the contact regions with the ground. Note that void particles have been removed, and the circles in white represent a solid with zero magnetization.

Under harmonic excitation, the optimized robot exhibits a cyclic walking motion driven by the asymmetry between its two limbs. During each gait cycle, when the magnetic field is directed upward (95 ms in Fig. 11E), the robot arches up to store elastic energy. Due to the imbalance in limb size and magnetic torque, the back limb dominates the motion, remaining the primary point of ground contact while lifting the forelimb off the ground. As the magnetic field reverses direction (98 ms in Fig. 11E), the robot curves downward. The back limb serves as an anchor, converting the released elastic energy into forward propulsion. Additionally, as the back limb rotates, its contact point shifts forward, further translating its clockwise torque into forward friction and reaction forces that enhance propulsion. In the next phase (101 ms in Fig. 11E), the forelimb makes contact while the back limb lifts off. As the magnetic field shifts upward again, it induces a clockwise torque in the grounded forelimb, generating propulsion and resistance forces that drive the robot forward. The robot completes one gait cycle (103 ms in Fig. 11E) and transitions into the next. Over the entire time frame, the robot advances by 2.01 mm. This case highlights how the co-design process integrates external stimulation, material properties, and structural configuration to achieve effective locomotion. By introducing asymmetry in size, force, and asynchrony between the back and forelimbs, the system optimally harnesses these factors to promote movement.

As shown in Fig. 12, the structure and materials co-evolve during optimization, but the history exhibits greater fluctuations than in the quasi-static MSCR cases. These are primarily due to the increasing binary projection sharpness ($\beta$) to handle the topological changes, which shrinks the design space and causes objective spikes after each update. A significant increase occurs between iterations 20 and 30, driven by a major topological change where the front section detaches to form the front limb and temporarily loses ground contact. This disrupts balance and paralyzes locomotion. However, the algorithm quickly recovers by elongating the limb to restore contact, enabling smooth convergence toward the final walking mode.

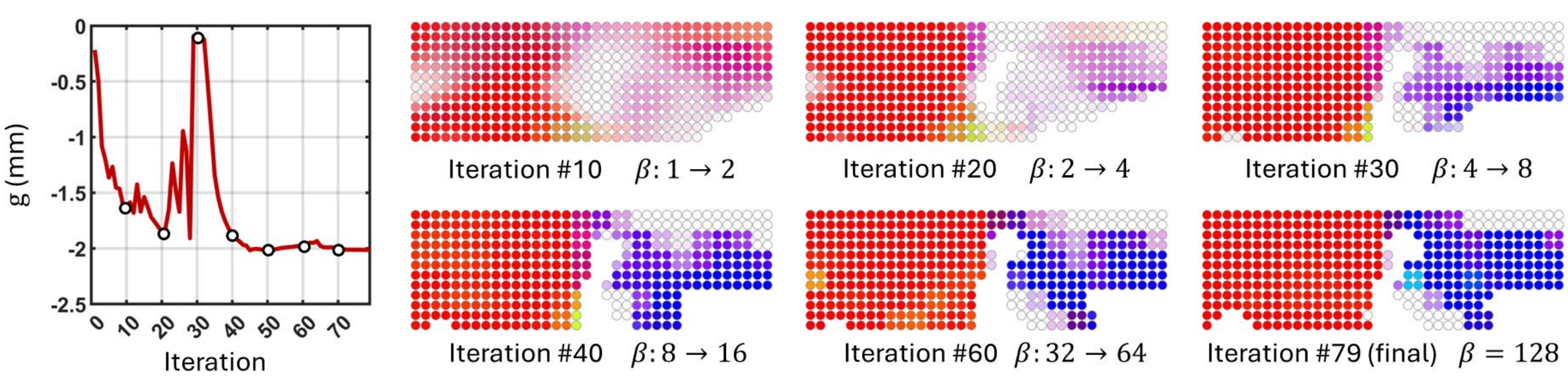

Fig. 12. Objective function history (left) and intermediate designs (right) for the first walker shown in Fig. 11. Dots on the objective function curves indicate update points for the binary projection sharpness parameter $\beta$.

In the second case (Fig. 13), where the structure has a smaller length-to-height ratio, the two-limb configuration appears again. Compared to the previous case, the limbs are more balanced in size. However, the magnetization in the back limb remains mostly perpendicular to the optimized magnetic field, while the forelimb is more aligned. As a result, while the imbalance is less pronounced than in the previous case, the back limb still generates a larger magnetic torque.

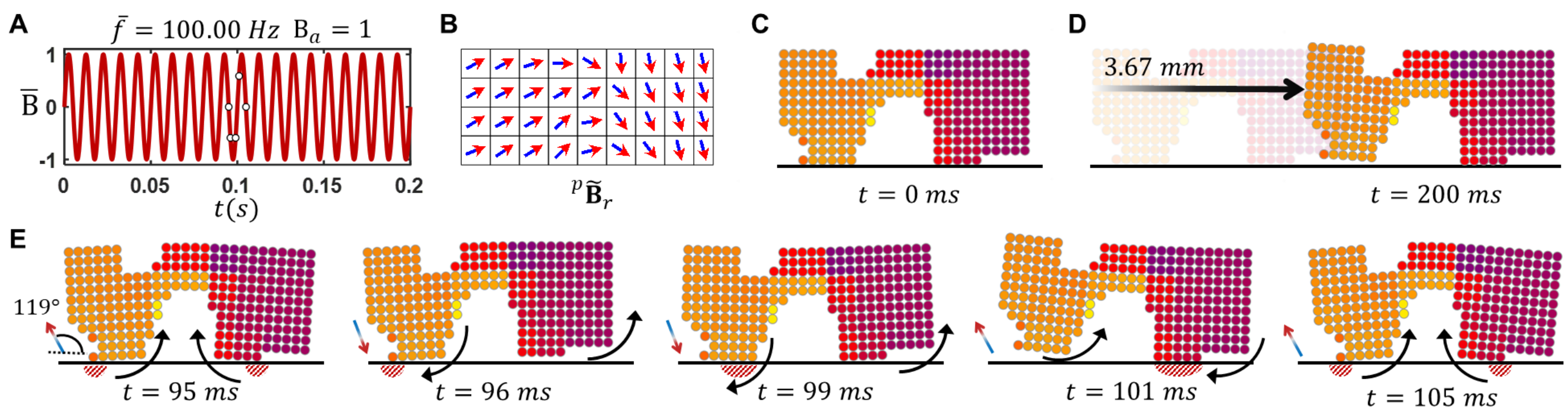

Fig. 13. Design results for the second walker. (A) Normalized magnitude of the optimized applied magnetic field. (B) Optimized magnetization. (C) Starting state, with the same color scheme as in Fig. 3B(ii). (D) Ending state, where the transparent shape shows the starting state. The arrow marks the total moving distance. (E) Snapshots of a locomotion cycle, corresponding to the dots in Fig. 11A.

The robot follows a similar gait cycle: first arching upward to store elastic energy (95 ms), then anchoring the back limb and releasing the stored energy to generate forward motion (96-99 ms). The back limb bends clockwise, providing forward propulsion, while the forelimb bends counterclockwise, contributing additional forward momentum. As the forelimb makes contact and the back limb lifts (when the field flips again), the grounded forelimb converts its clockwise bending into forward friction, while the back limb's counterclockwise bending continues to propel the robot forward (101 ms). This completes one gait cycle, allowing the robot to transition into the next. Due to the smaller length-to-height ratio and the more balanced limb sizes, the robot appears to store more elastic energy during the arching phase, as indicated by its greater curvature. Consequently, it moves forward a longer distance by 3.67 mm over the entire time frame, nearly its full body length.

For the second walker, the optimization history (Fig. 14) is noticeably smoother and converges more quickly than that of the first walker. This may be due to its shorter body length, which provides greater structural integrity and lower compliance, allowing the co-evolution of materials and structure to proceed without major disruptions during limb formation. Nevertheless, all the walker designs in this section required around 80 iterations to converge, with each iteration taking 16 to 17 seconds, resulting in a total optimization time of approximately 20 minutes. Despite using particle-wise topology variables, which significantly increase the design space's dimensionality, our method still demonstrated excellent robustness, scalability, and efficiency.

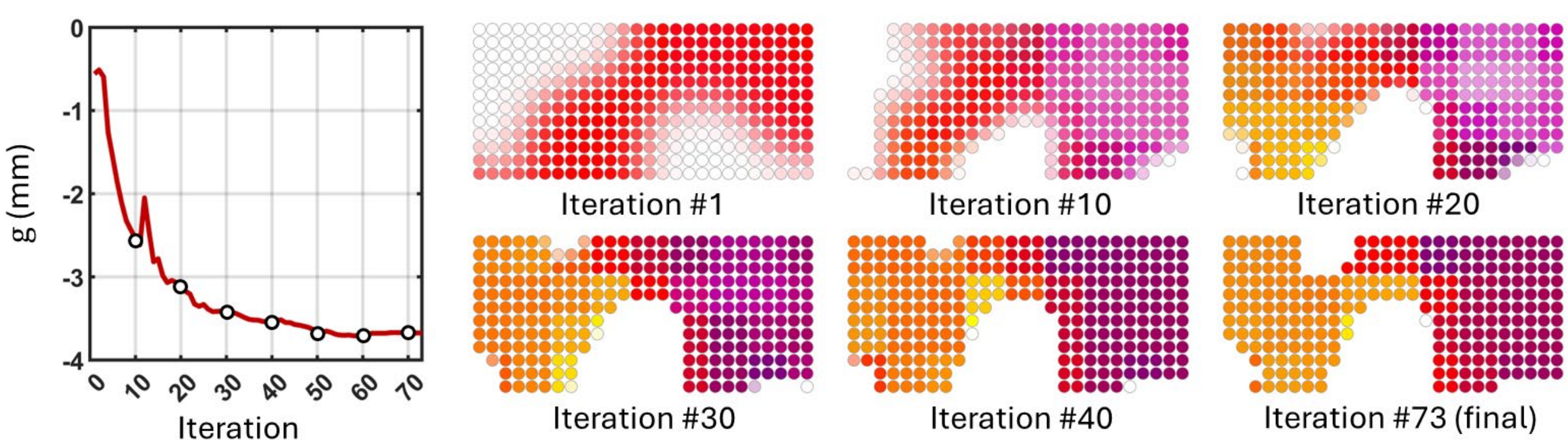

Fig. 14. Objective function history and intermediate designs for the second walker shown in Fig. 13.

Interestingly, the two optimized walkers exhibit two limb configurations similar to those found in artificial life or virtual creatures for comparable locomotion tasks [89-91]. However, unlike those approaches, which focus on materials with prescribed cyclic volumetric expansion and use evolutionary algorithms, our method employs magnetic torques with programmable stimuli and a more efficient gradient-based optimization. Despite fundamental differences, the emergence of similar structures suggests that the two-limb configuration may be optimal for this task, as also observed in nature [92]. This further highlights the capability of our method to reliably uncover effective design solutions.

### 4.3 Dynamic designs for 3D magnetic soft robot locomotion

Our method also accommodates 3D cases. As an example, we focus on the design of a slender crawler with dimensions of $10 \times 0.8 \times 2$ mm under a non-slip friction condition (Fig. 15A). The objective is to maximize its displacement along the x-axis by co-designing the magnetization orientation and external magnetic field, both constrained in the x-y plane. All the other settings remain the same as those in the previous 2D locomotion cases. The crawler is divided into 10 design regions along the x-axis. The optimization converged in 68 iterations, with each iteration taking approximately 10 seconds, resulting in a total optimization time of around 12 minutes (Fig. 16). The optimized design produces crawling motion similar to the 2D case shown in Fig. 10C. In this design, the tail section has magnetization almost perpendicular to the external magnetic field, which, under harmonic excitation, flaps cyclically to generate waves that propagate forward, driving the crawler (Fig. 15B-C). This design achieves a displacement of 0.7 mm in 0.2s, corresponding to a speed of 3.5 mm/s.

The flexible particle representation allows for the design of robots in complex 3D shapes. In Fig. 17A, we consider a cylindrical crawler, dividing it into a $5 \times 2$ grid of design regions in the x-z plane, as shown in Fig. 17B. We again co-design the material magnetization and applied magnetic field, this time allowing the magnetization to vary freely in 3D. As shown in Fig. 16, the optimization converged in 102 iterations, with each iteration taking approximately 7.5 seconds, leading to a total optimization time of around 13 minutes.

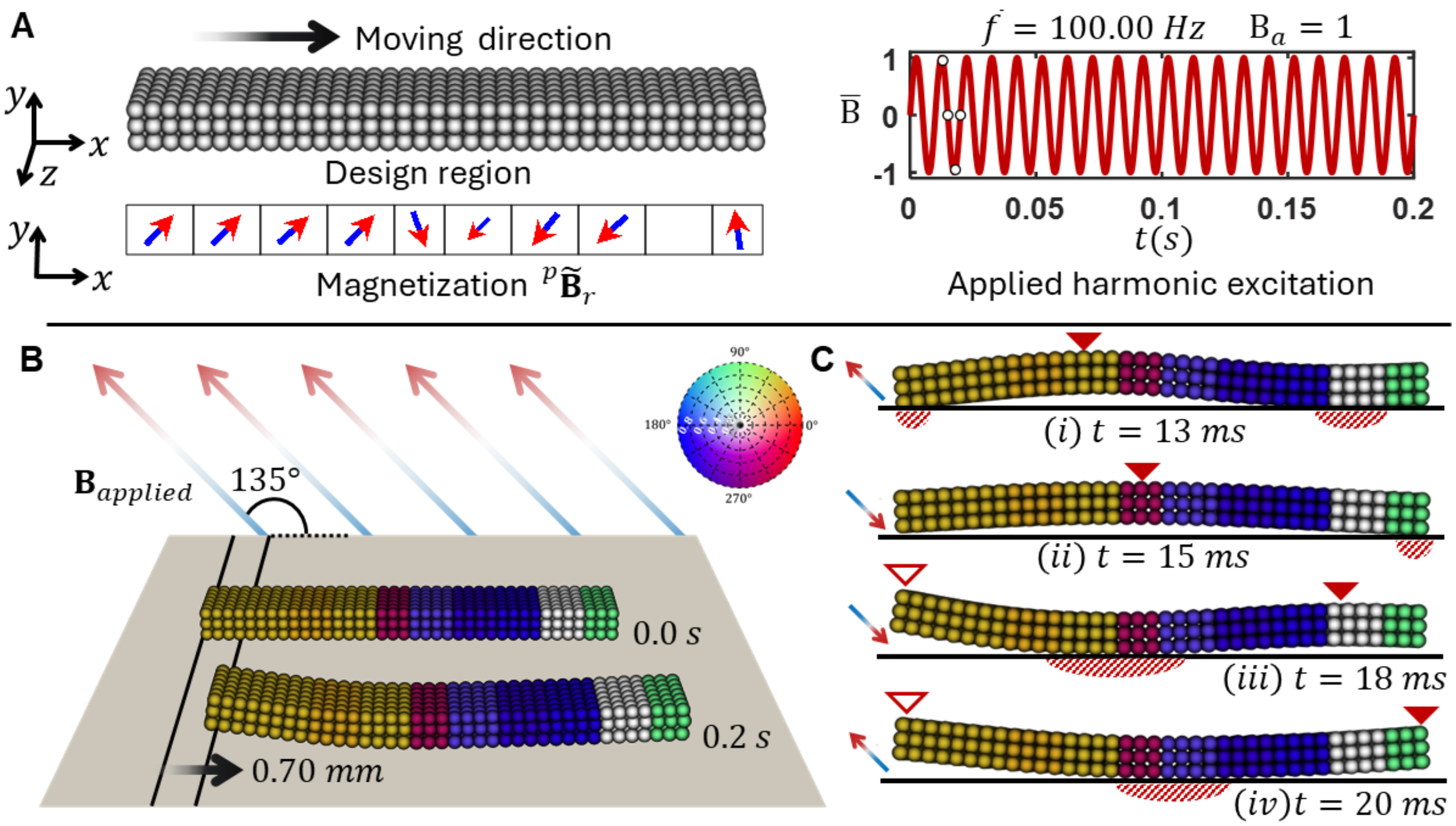

Fig. 15. Design results for the 3D slender crawler. (A) Particle representation (top-left), optimized magnetization in each design region (bottom-left), and optimized applied harmonic excitations (right). (B) Initial and final states of the designed crawler. Since the magnetization orientation is constrained within the x-y plane, the color coding represents the x-y components. (C) Snapshots of a locomotion cycle, corresponding to the dots in the excitation curve in Fig. 15A. The orientation of the applied magnetic field is shown in the lower left. Void triangles mark the formation of the hump, while solid triangles indicate the position of the formed hump. Patterned shades represent the contact regions with the ground.

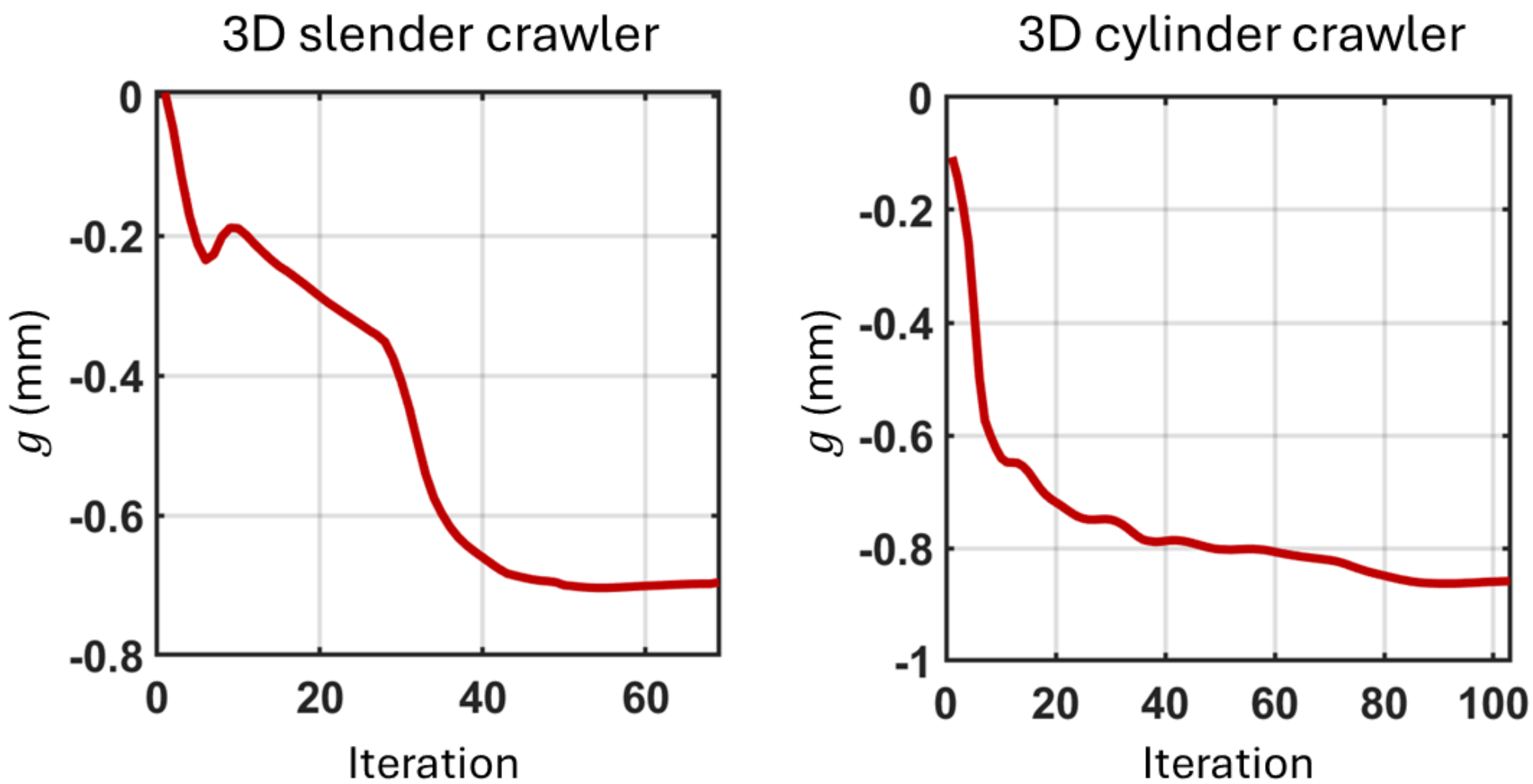

Fig. 16. Optimization history of the 3D slender crawler and the 3D cylindrical crawler.

From the optimized results in Fig. 17B, we observe that the magnetization in the back half (along the x-axis) primarily lies in the x-y plane and is perpendicular to the external magnetic field, enabling large flapping motions that generate driving forces. In contrast, the front half produces bending toques with an inverse orientation to the back half. The front half also includes magnetization components along the z-axis that promote bending around the x-axis. This asymmetry in magnetic torques between the back and front results in a crawling motion similar to the previous case, where a hump wave propagates from the back to the front (Fig. 17D, z-view). However, the crawler also adopts a saddle-like shape, exhibiting inverse curvatures along the x- and z-directions (as seen in the z- and x-views in Fig. 17D). Consequently, the contact regions continuously shift between the back and front, as well as the sides and center. This optimized design achieves a displacement of 0.86 mm in 0.2 s, corresponding to a speed of 4.3 mm/s.

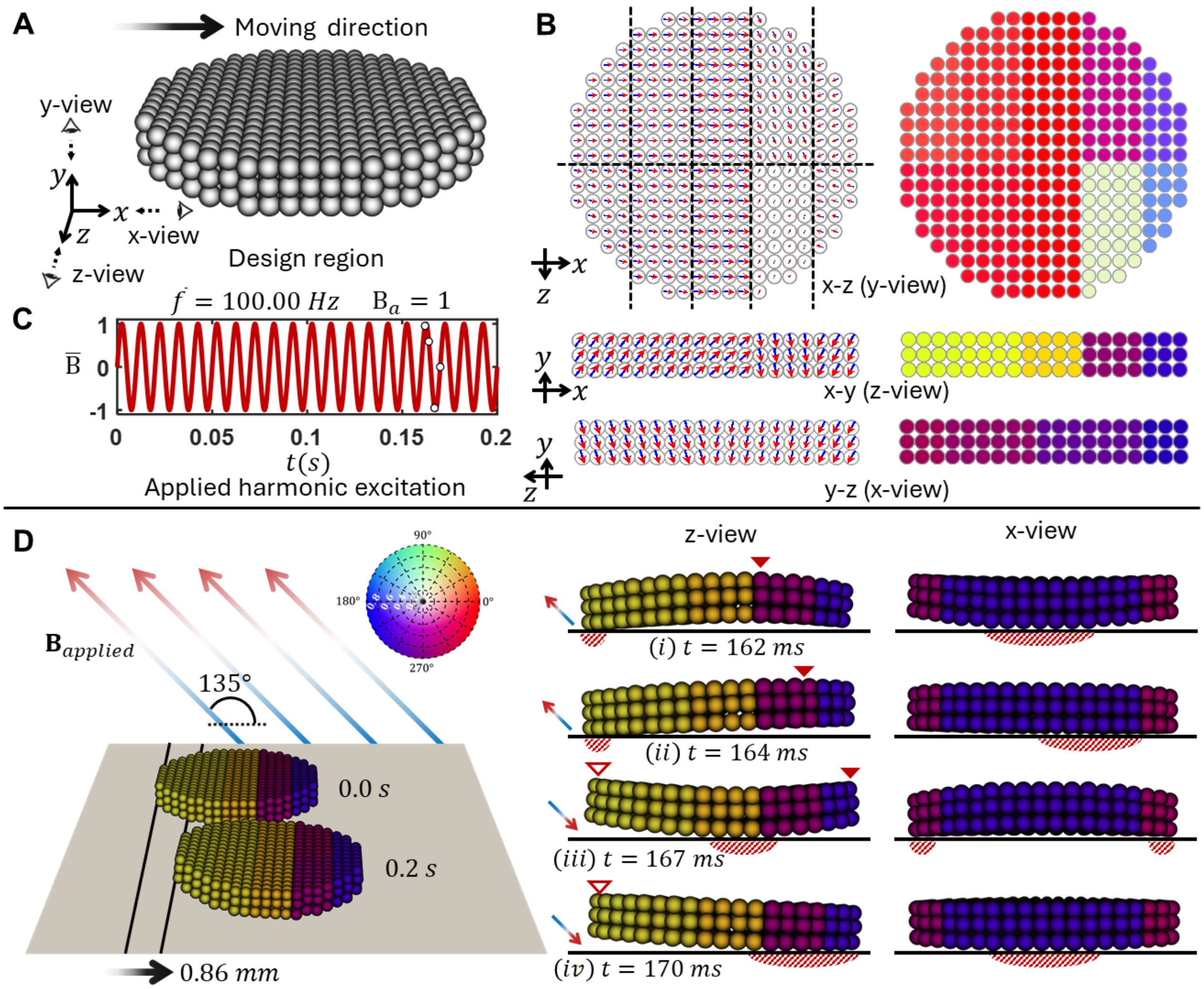

Fig. 17. Design results for the 3D cylinder crawler. (A) Particle representation with different views labeled alongside the coordinate axes. (B) Optimized magnetization shown in both arrow representation (left) and color visualization (right), displaying the projected components in different views. The dashed lines show the division of the design regions. (C) Optimized applied harmonic excitations. (D) Simulation of the locomotion: initial and final states of the designed crawler (left), snapshots of a locomotion cycle in the z-view (center column), and x-view (right column), corresponding to the dots in the excitation curve in Fig. 17C. For clarity, the color coding represents only the x-y components. The orientation of the applied magnetic field is shown in the lower left. Void triangles mark the formation of the hump, while solid triangles indicate the position of the formed hump. Patterned shading represents the contact regions with the ground.

## 5. Conclusions

We present an inverse co-design framework that simultaneously optimizes the structure, material magnetization distribution, and time-varying stimuli for magnetic soft robots. To accomplish this, we developed a unified design representation that integrates structure, materials, and stimuli together. These mixed-type variables are coupled through an interpolated magneto-elastic constitutive model, which is integrated into MPM for dynamic simulation. The GPU-accelerated MPM simulation ensures efficiency, and automatic differentiation provides sensitivity values for topology optimization. This enables materials, structures, and stimuli to co-evolve iteratively toward optimal designs. Unlike previous methods that focus on partial designs under static conditions, our framework supports co-design for both static and dynamic processes, accommodating large deformations and dynamic contacts in 2D and 3D. This flexibility is demonstrated in the design of magnetic soft robots capable of multi-task shape morphing and locomotion. In all cases, the method exhibits great efficiency, with the entire design process completed in minutes. Despite the complexity of the design space and coupled physics, optimized designs were achieved without relying on human intuition or expertise, rapidly adapting to new design tasks.

Our approach marks a fundamental step toward the automatic design of diverse magnetic robotic systems. Meanwhile, we would like to highlight several limitations and future directions of this work in terms of design representation, modeling, validation, and applications. For design representation, while this work focuses on harmonic excitations, it can be readily extended to incorporate more diverse signal forms and time-varying magnetic field orientations. On the modeling side, since our emphasis is on the co-design framework, we adopt a simplified magnetoelastic model based on some idealized assumptions [9]. Future work will focus on integrating more advanced theoretical developments that address complex real-world conditions such as viscoelasticity [12, 25, 26, 93] beyond the simple damping used in the current study, interaction with surrounding air and local neighboring components [73, 75, 82], deformation involving nonnegligible stretching [25, 26, 73-75], and more realistic contact and friction models [94]. This remains an active and fast-evolving area that requires joint theoretical and experimental efforts to bridge the gap between simulation and real-world physical behavior. Additionally, though we assumed a uniform magnetic field and ignored the effects of fast-varying fields, it is worth noting that non-uniform fields can be readily incorporated into the MPM framework. As discussed in Remark 1, these scenarios require replacing the $\mathbf{F}$-based model with the $\mathbf{R}$-based model to accurately account for magnetically induced stretching. This opens the possibility of using permanent magnets as stimuli, eliminating the need for bulky electromagnetic coil assemblies and offering greater control flexibility [32]. In terms of manufacturing and validation, although the modeling approach has been experimentally validated in various earlier studies [9, 17, 32, 34, 35, 67, 68, 76, 78-80], we do not explicitly account for fabrication or actuation constraints in this study. Specific material and manufacturing choices can also introduce additional complexities that are not fully captured by the current general simulation framework. While the primary goal of this study is to demonstrate the computational and design capabilities of the proposed framework, establishing a complete design–fabrication–testing pipeline, realizing optimized designs through advanced manufacturing techniques, and experimentally validating their performance are important directions for future research. Finally, for robotic applications, while we have focused on crawling and walking locomotion as proof-of-concept, future work will explore other moving modalities, such as rolling, spinning, jumping, and flipping [19, 95], and enable multi-modal magnetic robots [96]. Furthermore, MPM holds promise for designing robots that interact with other media, such as fluids, which would benefit biomedical applications [19, 84].

In summary, our work lays a computational foundation for the automated co-design of magnetic soft robots, enabling the effective exploration of the large design space and pushing the boundaries of soft robotics. While we focus primarily on magnetic soft robot design in this study, we highlight that our approach addresses several fundamental bottlenecks in the design of general responsive material systems. These include the co-design process, large deformations, time-varying and posture-dependent stimuli, and frictional contacts. Therefore, this framework holds great promise for the inverse design of responsive and soft materials in general. With its ability to handle complex physics, facilitate rapid optimization, and support diverse applications, this framework can accelerate the development of untethered robots for applications like minimally invasive procedures [8, 17], drug delivery [16], and beyond.

**Acknowledgments**

The author acknowledges the support from the Department of Mechanical Engineering at Carnegie Mellon University. The author also acknowledges Dr. Liu Wang, Dr. Lu Lu, and Dr. Shuai Wu for their insightful discussions.